\documentclass[journal,10pt]{IEEEtran} 
\usepackage{cite}   
\usepackage[pdftex]{graphicx}   
\usepackage{array}  
\usepackage[font={small},labelfont={bf}]{caption}
\usepackage{subcaption}
\usepackage{algorithm}
\usepackage[noend]{algpseudocode}
\usepackage{amsmath,amssymb,amsfonts}  
\usepackage{textcomp}
\usepackage{xcolor} 
\usepackage{caption}
\usepackage{mathtools}
\usepackage{bm}    
\usepackage{comment}
\usepackage{booktabs} 
\usepackage{multirow}
\usepackage{enumitem} 
\usepackage{stfloats} 
\usepackage{url} 
\begin{document}
\title{A Generic Machine Learning Framework for Radio Frequency Fingerprinting}
\author{Alex Hiles and Bashar I. Ahmad
\thanks{A. Hiles and B. I. Ahmad are with Applied Intelligence Labs, ITG, Defence, BAE Systems Digital Intelligence, Chelmsford, Essex, CM2 8HN, e-mail: bashar.ahmad@baesystems.com}}
\maketitle

\begin{abstract}
Fingerprinting radio frequency (RF) emitters typically involves finding unique characteristics that are featured in their received signal. These fingerprints are nuanced, but sufficiently detailed, motivating the pursuit of methods that can successfully extract them. The downstream task that requires the most meticulous RF fingerprinting (RFF) is known as specific emitter identification (SEI), which entails recognising each \textit{individual} transmitter. RFF and SEI have a long history, with numerous defence and civilian applications such as signal intelligence, electronic surveillance, physical-layer authentication of wireless devices, to name a few. In recent years, data-driven RFF approaches have become popular due to their ability to automatically learn intricate fingerprints. They generally deliver superior performance when compared to traditional RFF techniques that are often labour-intensive, inflexible, and only applicable to a particular emitter type or transmission scheme. In this paper, we present a generic and versatile machine learning (ML) framework for data-driven RFF with several popular downstream tasks such as SEI, data association (EDA) and RF emitter clustering (RFEC). It is emitter-type agnostic. We then demonstrate the introduced framework for several tasks using real RF datasets for spaceborne surveillance, signal intelligence and countering drones applications.
\end{abstract}

\section{Introduction}
\subsection{Background and Overview}
RFF involves building representations or profiles of RF emitters from distinct features that are present in their received transmissions. These can be utilised by a downstream task such as identifying each individual transmitter or distinguishing between emitter types (e.g. different radars, 5G sources, etc.) \cite{jagannath2022comprehensive, ren2025comprehensive}. 
In general, the sought emitters' characteristics can be intentional (e.g. the waveform parameters, modulation scheme, transmission timing or bandwidth, etc.), albeit impacted by the propagation channel effects (e.g. fading, dispersion, scintillation, etc.), or unintentional \cite{jagannath2022comprehensive, xie2024radio}. 
The latter often originate from inherent transmitter hardware imperfections (e.g. amplitude rise-time, ripple, phase distortion, nonlinearities in the processing chain, antenna responses, etc.) and/or are induced by manufacturing variabilities, 
including from individual hardware components \cite{jagannath2022comprehensive, al2024radio, ahmed2024comprehensive, ren2025comprehensive}. Spoofing or obfuscating the resultant (physical-layer) fingerprints is difficult. Overall, RFF entails defining and extracting salient features (intentional or not) that enable the recognition of the targeted emitters \cite{soltanieh2020review, al2024radio, ren2025comprehensive, abbas2024radio}, this is, RF fingerprint identification (RFFI).

In practice, data captures from a wide range of modulation schemes and transmission protocols are collected for coarse RFFI problems such as conventional automatic modulation or protocol identification \cite{huynh2021automatic, peng2019deep}. The primary focus in this paper is fingerprinting that permits solving more granular recognition tasks such as specific emitter identification. 
This is because the fingerprints in this case need to incorporate the most comprehensive and nuanced set of characteristics attributed to each individual emitter, typically exploiting unintentional features originating from hardware imperfections and/or manufacturing variabilities \cite{jagannath2022comprehensive, al2024radio, ren2025comprehensive}. For example, different instances of the same transmitter (i.e. same device model, make, transmission protocol, modulation scheme, etc.) can be uniquely identified with SEI; see the digital mobile radio and drones datasets in Section \ref{sec:experimentation}. Note that broader emitter classes (e.g. individual manufacturers) and even larger categories (e.g. modulation or protocol used), can be naturally derived and/or inform, implicitly or explicitly, by the detailed RFF for SEI. 

In general, there are two broad categories of RFF methods:
\begin{enumerate}
\item Traditional RFF with feature engineering and signal processing extraction techniques \cite{brik2008wireless,huang2017radio, danev2009physical, kennedy2008radio, ezuma2019micro, danev2009transient}. 
\item Data-driven, ML, typically deep learning (DL)-based,  RF fingerprinting \cite{sankhe2019oracle, robinson2020dilated, peng2019deep, shen2021radio, fu2023deep, wang2022few, wang2022specific, deng2023lightweight, wang2025sei}.
\end{enumerate}
Traditional RFF relies on manually defining the distinguishable features/characteristics that form the emitters' RF fingerprints such as evolution of the signal's frequency spectrum, central frequency offsets or its amplitude envelope \cite{jagannath2022comprehensive, wang2022few, deng2023lightweight, al2024radio}. Subsequently, a suitable signal processing (SP) algorithm is applied to obtain them from received transmissions. The fingerprint features are then ingested by a conventional classifier (inclusive of rule-based or classical machine learning methods such as support vector machines, decision trees, multi-layer perceptrons, etc.) for identification.

In general, traditional RFF methods can be effective, computationally efficient and their outputs fully explainable. However, they are inherently inflexible and usually specific to each emitter/signal type and a particular RFFI problem. They are labour-intensive requiring considerable RF specialists' effort to manually define the signal characteristics that constitute the RF fingerprint of each new type of emitter/signal and involve adapting/devising tailored SP techniques. This renders their development cycle slow and expensive. This can be problematic in the increasingly congested and contested electromagnetic environment (EME), where new emitters emerge regularly and communication technologies continuously evolve. Therefore, a more versatile and flexible approach to RFF might be desired.

On the other hand, data-driven RFF techniques leverage state-of-the-art (SOTA) ML, typically with deep neural networks (DNNs).~
 They have emerged in recent years as a promising RFF methodology due to the superior performance (especially SEI) they offer \cite{ding2018specific, peng2021survey, jagannath2022comprehensive, wang2022few, wang2022specific, deng2023lightweight, al2024radio}. DL-based RFF advantages are threefold. First, it automatically learns intricate fingerprint features from the data, for instance from in-phase/quadrature (I/Q) signals, whilst simultaneously performing one or more of several possible downstream tasks such as SEI, emitter data association and RF emitter clustering. 
This circumvents the aforementioned labour-intensive steps of the traditional RFF (i.e. defining fingerprint features and developing bespoke SP algorithms). Second, data-driven fingerprinting enables generic, multi-purpose and emitter-type agnostic, fingerprinting. It can be applied to various emitters and signal types including new, previously unseen emitters (with a suitably developed end-to-end ML pipeline). 
Third, ML-based RFF provides additional robustness against adverse channel conditions and interference \cite{jagannath2022comprehensive, wang2022few, al2024radio}. This facilitates emitter identification at lower signal-to-noise ratios (SNRs) compared to competing approaches, including those that reply on fully or partially demodulating-decoding the received signals from (non-)cooperative transmitters. 

\begin{figure}[!t]
  \centering
  \includegraphics[width=\linewidth]{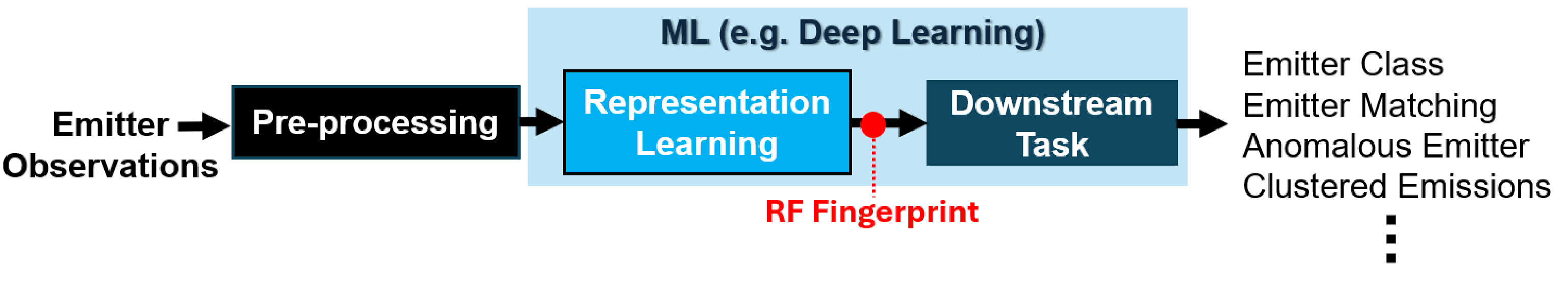}
  \vspace{-5mm}
  \caption{Block diagram of ML-based RFF-DT.}
  \label{mlpipeline}
  \end{figure}

A block diagram of ML-based RFF is depicted in Figure \ref{mlpipeline}. The representation learning module is where the RF fingerprints are learnt directly from the input data. This can be carried out under different ``levels of supervisions", 
referring to the relationship of the RF fingerprint representation learning block and the downstream task block that can be one or more of the following: (1) Identification of emitters, inclusive of SEI; 
(2) Comparing and matching multiple emitter emissions originating from the same emitter (e.g. EDA, typically solved using contrastive learning \cite{chen2020simple});
(3) Automatic grouping and building of RF fingerprint libraries for unknown emitters \cite{li2023class, smailes2025satiq};
(4) Anomaly detection based on learning RFF ``normalcy'' models, or a generative approach, e.g. variational auto-encoders (VAEs) \cite{zhang2024variational}; 
(5) Authentication (physical-layer) by comparing known fingerprints from those from received signals \cite{meng2025survey}.
Most importantly, a general RFF model can be learnt across numerous emitters or conditions, 
which can then be adapted or fine-tuned to a specific set of scenarios. This RFF downstream tasking (RRF-DT) has several applications such as for Signal Intelligence SIGINT (e.g. to detect-track threats), cyber security for wireless devices and networks (e.g. for internet of things IoT), Intelligence, Surveillance and Reconnaissance (ISR) and others \cite{jagannath2022comprehensive, xie2024radio, al2024radio, ren2025comprehensive, wang2022few}. 

ML-based RFF-DT, therefore, not only offers a faster development and iterative testing-refinement cycle, but also SOTA
performance. ~
This comes at the cost of needing to have access to sufficient representative data from the targeted emitters, with a certain level of ground-truth labels (i.e. supervision, if any) dependent on the the downstream task(s) \cite{jian2020deep, yang2021specific}. Nevertheless, few-shot learning methods permit adapting a model trained on data from \textit{similar} types of emitter(s) to those under investigation from a modestly small number of example RF captures \cite{wang2022few, jagannath2022comprehensive, al2024radio}. As is common in adopting data-driven, DL, methods there is a risk that the model does not learn the sought RF fingerprint (e.g. hardware-imperfections-related for SEI). 
Instead, it could latch onto the noise or EME profile or transmission pattern (e.g. frequency hopping) in the available data. To ensure mitigating such overfitting effects, it is crucial to employ appropriate best practices in ML/DL \cite{goodfellow2016deep} such as cross validation, model simplification, data augmentation with  synthetic examples, loss function regularisation, early stopping criteria for optimisers, ensemble methods, etc.

\subsection{Contributions}
The main contributions of this paper are:
\begin{itemize}
  \item A generic (mathematical) framework for ML-based RFF is introduced, providing a basis for treating various downstream tasks.
  \item A number of key individual RFF-related tasks (namely SEI, EDA and RFEC) are then formulated within this framework and solved using a variety of real RF datasets from automatic identification system (AIS) as captured by spaceborne receivers, digital mobile radios (DMRs) and drone RF links. 
  \item Performance of the ML-based RFF-DT is demonstrated with several model architectures; this is in the context of a number of use cases such as for ISR, SIGINT  and counter unmanned aerial systems (C-UAS).
\end{itemize}
This is preceded by a brief overview of SOTA ML-based RFF, encompassing input representations, challenges in closed set and open set problems, as well as mitigating the impacts of channel conditions and/or receivers responses.  
\subsection{Paper Layout}
The remainder of the paper is organised as follows. In Section \ref{sec:relatedwork}, a concise summary of ML-based RFF and associated challenges are presented. Whilst Section \ref{sec:framework} outlines the overall RFF framework, individual downstream tasks are stated in Section \ref{sec:singletasks}. The various use cases and real RF datasets are described in Section \ref{sec:experimentation} and results are discussed in Section \ref{sec:results}. Finally, conclusions are drawn in Section \ref{sec:conclusions}. 
\section{Related Work} \label{sec:relatedwork}
A large corpus of research on ML-based RFF exists in the general wireless communications community, especially for IoT and cyber security applications \cite{soltanieh2020review, wu2018deep, wang2022few, abbas2024radio}. 
Various DL models have been adopted and tailored for RF fingerprinting and identification. For example, convolutional neural networks (CNNs) \cite{yu2019robust}, recurrent neural networks (RNNs) \cite{merchant2019enhanced}, 
long short-term memory (LSTMs) \cite{shen2021radio}, transformers \cite{vaswani2017attention}, generative adversarial networks (GANs) \cite{merchant2019securing}, auto-encoders (AEs) \cite{yu2019radio}, 
variational auto-encoders (VAEs) \cite{jiang2024radio}, amongst many others have all been applied towards RFF-based problems. CNNs in particular, including complex valued CNNs (CV-CNNs) \cite{kashani2024radio}, are amongst the most widely used model architectures. 
This is generally attributed to their wide availability across different ML libraries/tools and ability to capture temporal correlations in the processed RF signals. Next, we highlight a few key aspects associated with RFF inclusive of the nature of the downstream problem, input representations and impact of channel effects.
\subsection{Closed-set vs Open-set Tasks}
The various RFF-based downstream tasks, except those that use  unsupervised learning, can be viewed as being a:
\begin{enumerate}
\item {\bf{Closed-set downstream task (CS-DT)}}: constrained to a predefined finite set of emitters, for which there is existing labeled data/recordings.
\item {\bf{Open-set downstream task (OS-DT)}}: new, previously unseen (i.e. in model training), RF emitters  can be encountered, requiring suitable techniques to recognise their presence and address them appropriately (e.g. identify them if detected again, if relevant).
\end{enumerate}
The vast majority of research in ML-based RFF  are CS-DT based approaches. For OS-DT, new emitters with respect to a trained CS-DT model need to be identified (e.g. compared to the closed-set members) and their fingerprints learnt such that they can be recognised
in a future iteration of the CS-DT model \cite{xie2025novel}. There are several approaches which have been taken to address this issue, including class incremental learning \cite{li2025fscil, li2023class, li2017learning, zhu2021class}. 
These methods can be viewed in the context of anomaly detection for cyber security applications and/or detecting adverse channel conditions.   
\subsection{Input Representations}
Broadly speaking, ML-based RFF-DT methods can be categorised by their input:
\begin{enumerate}
\item {\bf{Time-domain signal}}: refers to the received I/Q data of the detected transmissions from the targeted emitter and/or their resultant components from empirical mode decomposition, variational mode decomposition, etc. CNNs and RNNs are a common choice for this data representation, including CV-CNNs and LSTMs. Directly ingesting the I/Q data can be highly desirable for a multi-purpose, versatile, RFF capability for various emitter types and scenarios. 
\item {\bf{Spectrogram or transform-domain}}: bi-spectrum analysis \cite{liu2021incremental} and short-time Fourier transformer (STFT) \cite{shen2022towards}, Hilbert-Huang \cite{lv2014automatic}, and wavelet transforms \cite{xie2018optimized} as well as histograms and data groupings fall into this class. The transform  representations (or information extracted from them) are then fed into a neural network, e.g. a CNN. Any transformation should preserve the relevant information in the (complex) time domain RF data to facilitate the pursued feature extraction capability of the DL algorithms.
\item {\bf{Constellation-based}}: relies on the differential constellation trace figure, often based on the modulation scheme employed by the emitter \cite{peng2019deep}. This modality is applicable to communication signals and the constellation traces typically serve as an input to the DL model (e.g. a multi-channel CNN). However, emitter classification from the constructed trace can be challenging in practice and is specific to an emitter type/modulation-scheme, i.e., leads to an inflexible RFF capability. \end{enumerate}
The time series inputs can include the transient and/or the steady-state segments of the received signal \cite{soltanieh2020review, wang2022few, deng2023lightweight}. The former is related to the start of the transmission (e.g. initial part of a burst inclusive of its rising edge) and often incorporates prominent unique distinguishable characteristics related to the emitter's hardware. Accurate detection of the start of  emissions can be difficult, especially in challenging EMEs. The steady-state segment follows from the transient phase and can include message/payload-related data, if relevant. It still nonetheless contains salient emitter features that enable SEI \cite{wang2022few, deng2023lightweight}. Processing a combination of transient and steady-state components of the transmission can often be the pragmatic approach in practice.
\subsection{Mitigating Channel and Receiver Response Effects}
In practice, the received signal can be substantially impacted by multipath, 
fading and noise from the propagation channel. 
Similarly, inherent differences amongst various deployed receivers (e.g. distributed on different satellites in a constellation) can induce a significant mismatch (e.g. distribution shift) between 
the training dataset and the signals that the ML RFF approach must handle when deployed. Naturally, these scenarios degrade the performance of any RFF-based model. 
Therefore, successful RFF-based models need to incorporate, where needed, model adaptation, domain adaptation, transfer learning and few-shot learning approaches. 
All of these approaches enable generalisable and robust ML-based RFF as well as SEI \cite{zhang2024domain, li2025non, wang2020radio, wang2025avoiding}. The aim of these approaches
is to improve the resilience of the model in light of certain environmental and sensing conditions. These methods have been tested and applied in numerous research papers on real and synthetic datasets that pertain to complex environments 
(e.g. dense urban or indoor settings) and for transmission frequencies which are particularly susceptible to channel effects, for example deep fading such as with ZigBee \cite{yu2019robust}, Bluetooth low energy \cite{ali2019assessment} and Wi-Fi \cite{hanna2022wisig}. 
\section{Problem Statement and  RFF Framework} \label{sec:framework}
We formally introduce here the framework for learning general-purpose RF fingerprint representations using ML.  
Let us assume that we have a dataset $\mathcal{D} = ({\bf{X}}, {\bf{Y}}) \in \mathcal{X} \times \mathcal{Y}$ where ${\bf{X}}$ and ${\bf{Y}}$ are sampled from the unknown joint probability distribution $\mathcal{P}_{\mathcal{X}\mathcal{Y}}$, where $\mathcal{X}$ is the RF feature space and
$\mathcal{Y}$ is the label space. 
We define a ${\it{task}}$ $\mathcal{T}$ to be composed of the dataset $\mathcal{D}$, a loss function $\mathcal{L}$ and model parameterisations of ${\bf{g}}_\phi$ and ${\bf{F}}_\theta$, 
where ${\bf{g}}_\phi$ is the task head, and ${\bf{F}}_\theta$ is a vector-valued RFF function that contains, or is equivalent to, the RF fingerprint head ${\bf{f}}_\theta: \mathcal{X} \mapsto \mathcal{Z}$, where $\mathcal{Z}$ is the RFF representation (i.e. embeddings) space. The latter function performs the representation learning as depicted in Figure \ref{mlpipeline} (i.e. produces the sought RF fingerprint from a given input RF data). 
In general, ${\bf{F}}_\theta: \mathcal{X} \times \cdots \times \mathcal{X} \mapsto [\mathcal{Z}]^{l}$, where $l$ denotes the number of RF fingerprint head projections that have been made and $[\mathcal{Z}]^l \triangleq \mathbb{R}^{d \times l}$. 
For the majority of tasks, such as SEI, $l=1$ and the vector-valued function ${\bf{F}}_\theta \equiv {\bf{f}}_\theta$ whereas for other tasks, e.g. EDA, $l > 1$. Without loss of generality,  ${\bf{g}}_\phi$ and ${\bf{F}}_\theta$, jointly or individually, can be ML (DNN) models with the associated (hyper)parameters (i.e. $\phi$ and $\theta$) depending on the pursued downstream task(s).      

The composite function ${\bf{h}}_{\xi}({\bf{x}}) \triangleq {\bf{g}}_{\phi}({\bf{F}}_{\theta}({\bf{x}}))$ projects the input data ${\bf{x}}\in {\bf{X}}$ to a prediction $\hat{{\bf{y}}}$, which is computed for every ${\bf{x}}\in {\bf{X}}$, forming a predictive labels set $\hat{{\bf{Y}}}=\{ \hat{ {\bf{y}}} | \hat{ {\bf{y}}} = {\bf{h}}_\xi({\bf{x}}), {\bf{x}}\in {\bf{X}}\}$. 
The loss function $\mathcal{L}$ measures the distance between any pairs of $( {\bf{y}}, \hat{{\bf{y}}}) \in {\bf{Y}} \times \hat{{\bf{Y}}}$ and is used to supervise the model training, and ${\bf{g}}_\phi: [\mathcal{Z}]^{l} \mapsto \mathcal{Y}$ 
is a mapping from $[\mathcal{Z}]^l$ to the label space $\mathcal{Y}$, where relevant.

Notationally, we represent the $k$th ${\it{task}}$ with superscripts (e.g. $t$th task dataset $\mathcal{D}^{(t)}$), where each task is assumed to be sampled from the probability distribution of tasks $\mathcal{P}_{\mathcal{T}}$ 
and $\mathcal{D}^{(t)}=({\bf{X}}^{(t)},{\bf{Y}}^{(t)}) = \{ ({\bf{x}}_i^{(t)}, y_i^{(t)}) \}_{i=1}^{n_t} \sim \mathcal{P}_{\mathcal{X}^{(t)}{\mathcal{Y}^{(t)}}}$. Therefore, we also have the $t$th task predictive labels set $\hat{{\bf{Y}}}^{(t)}$. 
Note that for some tasks, the parameter set $\phi$ is fixed since the function ${\bf{g}}$ is a deterministic function (e.g. distance measure) but for others is learnable.
For example, ${\bf{g}}_\phi$ maps the inferred lower-dimensional characteristic representation (i.e. RF fingerprint) to the classification label space for a SEI task, whereas for an EDA task, 
${\bf{g}}_\phi$ is a distance measure (e.g. cosine distance) between two RF input embeddings. Hence, $\phi$ is not strictly learnable whereas $\theta$ (i.e. the representation learning model parameters such as the weights in a DNN) is.

Within a multi-task RFF context, we first introduce the following optimisation problem:
\begin{equation}
\min_{\xi\in\Xi} \mathbb{E}_{t\sim \mathcal{P}_{\mathcal{T}}} \left[ \mathbb{E}_{  \mathcal{D}^{(t)} \sim \mathcal{P}_{\mathcal{X}^{(t)} \mathcal{Y}^{(t)} } }\mathcal{L}^{(t)} \left({\bf{h}}^{J, (t)}_{\xi_t}\left( {\bf{X}}^{(t)}\right), {\bf{Y}}^{(t)}\right) \right], \nonumber 
\end{equation}
where ${\bf{h}}_{\xi_t}^{J, (t)}( {\bf{X}}^{(t)}) = {\bf{g}}_{\phi_t}^{(t)}\left({\bf{F}}_{\theta}^{(t)}\left( {\bf{X}}^{(t)}  \right)\right)$, with superscript $J$ for $\theta$ that is jointly learnt across tasks,  
$\Xi = \Theta \times {\bf{\Phi}}$, $\Theta \equiv \mathbb{R}^d, d \in \mathbb{Z}^+$, ${\bf{\Phi}} = \Phi_1 \times \cdots \times \Phi_{|\mathcal{P}_{\mathcal{T}}|}$, $\Phi_{j}\equiv \mathbb{R}^{c_j}, c_j \in \mathbb{Z}^+$, 
with $\theta^* \in \Theta$, ${\boldsymbol{\phi}}^*=(\phi^*_1, \cdots, \phi^*_{|\mathcal{P}_{\mathcal{T}}|})$ with $\phi^*_j \in \Phi_j \; \forall j \in \{1, \cdots, |\mathcal{P}_{\mathcal{T}}|\}$. Upon solving the above optimisation task, 
we jointly learn the RF fingerprinting model parameters $\theta$ and the task-specific parameters $\phi_t$, for all $t$.

We define the parameters $\xi^*$ as an argument minimiser of the optimisation problem above. Intuitively speaking, it is expected that the resulting RFF head ${\bf{f}}_{\theta^*}$ has an increasingly richer 
fingerprint representation when jointly learning $\theta$ across an increasing amount of tasks that require RF fingerprint knowledge. It is assumed that the functional form of the RF fingerprint head is chosen to be the same across all tasks. In other words, ${\bf{F}}_{\theta}^{(t)}={\bf{F}}_\theta$
where ${\bf{F}}_\theta$ is chosen appropriately. 

Alternatively, we can learn across all tasks but have independent parameter sets $\theta_t$ for each task $t$ and aggregate them post optimisation. Mathematically speaking,
\begin{equation}
\min_{\xi\in\Xi}\mathbb{E}_{t\sim \mathcal{P}_{\mathcal{T}}} \left[ \mathbb{E}_{\mathcal{D}^{(t)} \sim\mathcal{P}_{\mathcal{X}^{(t)}\mathcal{Y}^{(t)}}} \mathcal{L}^{(t)} \left({\bf{h}}_{\xi_{t}}^{I, (t)}( {\bf{X}} ^{(t)} ), {\bf{Y}}^{(t)} \right) \right],\nonumber 
\end{equation}
where ${\bf{h}}_{\xi_{t}}^{I, (t)}( {\bf{X}} ^{(t)} ) = {\bf{g}}_{\phi_t}^{(t)}\left( {\bf{F}}_{\theta_t}^{(t)}\left({\bf{X}}^{(t)} \right)   \right)$, with $I$ denoting that an independent parameter set $\theta_t$ is being learnt for each task. 
An aggregation function $a : \Theta_1 \times \cdots \times \Theta_{\lvert \mathcal{P}_{\mathcal{T}} \rvert} \mapsto \mathbb{R}^d$ can then be used to combine the parameters into one parameter set $\theta^*$. 
In other words, $\theta^* = a({\boldsymbol{\theta}}^*).$ One possible composition of the aggregation function $a$ could be $a({\bf{x}}) = \frac{1}{n}\sum_{i=1}^n x_i$, for example.

In the discrete setting, we assume that $m$ tasks exists, where the $t$th task has a dataset $\mathcal{D}^{(t)}$, with each data point ${\bf{z}} \in \mathcal{D}^{(t)}$ being independent identically distributed (IID) 
and assumed to be drawn from an unknown joint probability distribution $\mathcal{P}_{\mathcal{X}^{(t)}\mathcal{Y}^{(t)}}$. This results in the following optimisation problems:
\begin{equation}
\boldsymbol{\xi}_{S, J}^*= \mbox{arg}\min_{\xi\in\Xi} \sum_{t=1}^m \frac{\alpha_t}{n_t} \sum_{j=1}^{n_t} \mathcal{L}^{(t)} \left({\bf{h}}^{J, (t)}_{\xi_t} \left({\bf{x}}^{(t)}_{j}\right), {\bf{y}}^{(t)}_{j} \right), \nonumber 
\end{equation}
\begin{equation}
\boldsymbol{\xi}_{S, I}^* = \mbox{arg}\min_{\xi\in\Xi} \sum_{t=1}^{m} \frac{\alpha_t}{n_t} \sum_{j=1}^{n_t} \mathcal{L}^{(t)} \left({\bf{h}}^{I, (t)}_{\xi_t}\left(  {\bf{x}}^{(t)}_{j}\right), {\bf{y}}^{(t)}_{j} \right), \nonumber 
\end{equation}
where the argument minimisers ${\boldsymbol{\xi}}_{S, J}^*= (\boldsymbol{\phi}_{S, J}^*, {\theta}_{S, J}^*)=(\phi_1^*, \cdots, \phi^*_{n}, \theta^*)$ 
  and  $\boldsymbol{\xi}_{S, I}^* = ({\boldsymbol{\phi}}_{S, I}^* ,\boldsymbol{\theta}_{S, I}^*) =(\phi_1^*, \cdots, \phi^*_{m}, \theta^*_1, \cdots, \theta^*_m)$ for the joint and individual (parameter set for each task) optimisation tasks respectively, and 
$\alpha_t$ denotes the weighting of the $t$th task with $\sum_{t=1}^n \alpha_t = 1$. A standard choice for this is equal weighting across tasks (e.g. $\alpha_t = 1/n$ for all $t$), though this formulation allows for this to be selected according to preference. 
In the same manner as previously discussed, we can perform aggregation to find $\theta^* = a({\boldsymbol{\theta}}^*) = a(\theta_1^*, \cdots, \theta_m^*) $.

The resulting parameters found from solving either of these optimisation problems are a function of the RFF-based tasks, data quality, loss function and model architecture. 
In practice, whilst $\mathcal{P}_{\mathcal{T}}$ is unknown, it is assumed that common tasks such as SEI, EDA and RFEC, amongst others, are the 
modes of this distribution and approximate the expectation appropriately. Therefore, it is likely that approximate solutions to this optimisation problem can be found by jointly considering many popular RF fingerprint-based tasks.

The joint optimisation problems introduced here are analogous to multi-modal joint embedding learning techniques \cite{wang2023image,radford2021learning, alayrac2022flamingo}, 
though here the intuition is on solving different tasks in the multi-task learning (MTL) sense \cite{sener2018multi, zhang2021survey,caruana1997multitask}, rather than enriching the RF representation via multi-modal data. 
Both formulations are multi-objective optimisation problems that serve multi-task learning \cite{sener2018multi} and are a generic approach towards learning RF fingerprint representation spaces via data-driven methods.
\section{Single Task RF Fingerprint Learning} \label{sec:singletasks}
We now present a few popular RF fingerprint-based tasks, each considered individually.
\subsection{Closed-Set Specific Emitter Identification}\label{sec:singletask_CSSEI}
The first task we define is closed-set SEI (CS-SEI). The objective is to learn an accurate mapping from emitter observations (e.g. I/Q samples) to the correct emitter label.
Assuming that $\mathcal{P}_{{\mathcal{X}}^{(i)}{\mathcal{Y}}^{(i)}}$ is the joint probability distribution of the SEI feature space $\mathcal{X}^{(i)}$ and the emitter label space $\mathcal{Y}^{(i)}$,
where the superscript $i$ denotes identification task. We pose the following optimisation problem to learn a mapping ${\bf{h}}_{\xi}$ from ${\mathcal{X}}^{(i)}$ to ${\mathcal{Y}}^{(i)}$:
\begin{eqnarray}
\xi^* = \mbox{arg}\min_{\xi \in \Xi} \mathbb{E}_{( {\bf{x}} , y)\sim \mathcal{P}_{\mathcal{X}^{(i)}\mathcal{Y}^{(i)}}}\mathcal{L}^{(i)}\left({\bf{h}}^{(i)}_\xi\left( {\bf{x}} \right) , y\right). \nonumber
\end{eqnarray}
This translates to finding a suitable parameter set pair $\xi^* = (\phi^*, \theta^*)$ such that ${\bf{h}}_{\xi^*}( {\bf{x}})={\bf{g}}_{\phi^*}({\bf{F}}_{\theta^*}( {\bf{x}}))\approx y \; \forall ({\bf{x}}, y) \in \mathcal{X}^{(i)} \times \mathcal{Y}^{(i)}$, noting here that for CS-SEI ${\bf{F}}_{\theta^*} \equiv {\bf{f}}_{\theta^*}$.

Since the joint probability distribution is unknown in practice, we cannot directly solve this optimisation problem in its exact form. Instead, we minimise the empirical error, replacing the joint distribution with the sampled empirical data.
We assume a dataset $\mathcal{D}^{(i)}= ({\bf{X}}^{(i)}, {\bf{Y}}^{(i)})$ is sampled from $\mathcal{P}_{{\mathcal{X}}^{(i)}{\mathcal{Y}}^{(i)}}$, 
which we expand as $\mathcal{D}^{(i)}=\{({\bf{x}}^{(i)}_k, y^{(i)}_k)\}_{k=1}^{n_i}$ where ${\bf{x}}^{(i)}_k\in\mathbb{R}^{d^{(i)}}$, $y_k\in \{1, 2, \cdots, \mathcal{V}\}$, $n_i$ is the number of samples for the identification task and $\mathcal{V}$ is the number of unique emitters in the considered closed-set. 
We assume that each data point ${\bf{z}}=({\bf{x}}, y)\in \mathcal{D}^{(i)}$ is IID sampled from $\mathcal{P}_{{\mathcal{X}}^{(i)}{\mathcal{Y}}^{(i)}  }$. The CS-SEI optimisation problem is defined as:
\begin{eqnarray}
\xi_{em}^* = \mbox{arg}\min_{\xi \in \Xi} \mathbb{E}_{({\bf{x}}, y)\sim \mathcal{D}^{(i)}}\mathcal{L}^{(i)}\left({\bf{h}}^{(i)}_\xi\left( {\bf{x}} \right), y\right), \nonumber
\end{eqnarray}
where the subscript $em$ denotes empirical.
\subsection{RFF-based Emitter Data Association} \label{sec:singletask_EDA}
The objective in this task is to learn a model that can determine whether a pair of emitter observations originate from the same emitter or not. 
In other words, we are interested in finding a function ${\bf{h}}_\xi: {\mathcal{X}}^{(a)} \mapsto \mathcal{Y}^{(a)}= \{0,1\}$, where ${\mathcal{X}}^{(a)}$ is the EDA feature space, 
and labels 0 and 1 denotes no match and match, respectively. In this case, the model prediction $\hat{y}={\bf{h}}^{(a)}_\xi ({\bf{x}})$ is parameterised as follows:
\begin{equation}
{\bf{h}}^{(a)}_\xi ({\bf{x}}) = {\bf{g}}^{(a)}_\phi\left({\bf{F}}^{(a)}_\theta \left( {\bf{x}} \right) \right) = {\bf{g}}^{(a)}_\phi \left({\bf{f}}^{(a)}_\theta \left({\bf{x}}_1\right), {\bf{f}}^{(a)}_\theta \left({\bf{x}}_2\right) \right), \nonumber
\end{equation}
where ${\bf{g}}^{(a)}_\phi$ here is a distance metric (e.g. Euclidean, cosine distance, etc.), ${\bf{F}}^{(a)}_\theta$ is a RFF function containing two separate projections of emitter data using the underlying RF fingerprint head ${\bf{f}}^{(a)}_\theta$
and ${\bf{x}}=({\bf{x}}_1, {\bf{x}}_2)\in {\mathcal{X}}^{(a)}\equiv \mathbb{R}^{d^{(a)}  }   \times \mathbb{R}^{d ^{(a)}}$ where $d^{(a)}$ is the dimensionality of an emitter's feature space for the EDA task. This results in the following optimisation problem: 
\begin{eqnarray}
\xi^* = \mbox{arg}\min_{\xi \in \Xi}\mathbb{E}_{({\bf{x}}, y) \sim \mathcal{P}_{\mathcal{X}^{(a)}Y^{(a)}}}\mathcal{L}^{(a)}  \left({\bf{h}}_\xi^{(a)}\left({\bf{x}}\right), y \right) ,\nonumber
\end{eqnarray} 
where $\mathcal{P}_{\mathcal{X}^{(a)}{Y}^{(a)}}$ is the joint distribution of the RF feature space $\mathcal{X}^{(a)}$ and the task label space $\mathcal{Y}^{(a)}$; $\mathcal{L}^{(a)}$ is the EDA loss function. 

Like CS-SEI, this optimisation problem is solved in practice by collecting empirical datasets, assumed to be sampled from the typically unknown joint probability distribution. Without loss of generality, it is assumed here that an empirical dataset $\mathcal{D}^{(a)}$ is transformed from the CS-SEI dataset $D^{(i)}$. 
In other words, $\mathcal{D}^{(a)} = \mathcal{M}(\mathcal{D}^{(i)}) = \{ ( {\bf{x}}_{1,k}^{(a)}, {\bf{x}}_{2,k}^{(a)}, y^{(a)}_k) \}_{k=1}^{n_{a}}$, where $n_{a}$ is the number of samples in the dataset, $\mathcal{M}$ is a data transform operator such as the one outlined in Appendix A, $y^{(a)}_k \in \{0,1\}$ and $({\bf{x}}_{1,k}^{(a)}, {\bf{x}}_{2, k}^{(a)}) \in {\bf{X}}^{(a)}\equiv \mathbb{R}^{d^{(a)}} \times \mathbb{R}^{d^{(a)}}$.
This results in:
\begin{eqnarray}
\xi^*_{em} = \mbox{arg}\min_{\xi \in \Xi} \mathbb{E}_{({\bf{x}}_1, {\bf{x}}_2, y) \sim \mathcal{D}^{(a)}}\mathcal{L}^{(a)}\left(\hat{y}, y\right); \nonumber \\ 
           = \mbox{arg}\min_{\xi \in \Xi} \frac{1}{n_a}\sum_{k=1}^{n_{a}} \mathcal{L}^{(a)}\left( \hat{y_k}, y_k \right). \nonumber 
\end{eqnarray}
For the data transformation method presented in Appendix A, $n_a$ is selected according to a sampling strategy because for large $n_i$ the combinatorics become difficult to manage. We also note here that collecting EDA-specific empirical datasets is also possible. 
\subsection{Unsupervised Learning for RFEC}\label{sec:singletask_RFEC}
We recall that RFEC aims to determine and group the received RF transmissions based on the their sources, where the number and identity of the encountered emitters are unknown \textit{a priori}. Hence no labels are available and unsupervised learning methods are applied. The obtained ``clustering`` and distinctions between the observed groups of emitters is expected to be driven by the automatic extraction of RF fingerprints \cite{morge2022rf, jagannath2022comprehensive, ren2025comprehensive}. Common methods for learning data structures and hidden patterns without labels are clustering approaches such as, K-means, Gaussian mixture models, hierarchical clustering, DBSCAN \cite{ester1996density} and many others. These can be applied on the output of a dimensionality reduction technique, such as principal component analysis (PCA), or low-dimensional representations of the data inclusive within an auto-encoder architecture \cite{goodfellow2016deep}. The latter is adopted here where we describe an AE-based RFEC.

Autoencoders typically involve projecting the original data into a lower dimensional representation space, where the encoded data is then fed through a decoder to reconstruct the original input. 
If the reconstruction error is small, then the encoded representation is assumed to contain the important salient information of the original data. 
In this setting, the optimisation problem is:
\begin{eqnarray}
\xi^* = \mbox{arg}\min_{\xi \in \Xi} \mathbb{E}_{ {\bf{x}} \sim \mathcal{P}_{\mathcal{X}^{(r)}}}\mathcal{L}^{(r)}\left(\hat{{\bf{x}}}, {\bf{x}}\right), \nonumber 
\end{eqnarray}
where ${\bf{\hat{x}}}= {\bf{g}}^{(r)}_\phi({\bf{F}}^{(r)}_\theta ({\bf{x}}) ) $, $\mathcal{P}_{{\mathcal{X}}^{(r)}}$ is the probability distribution of the emitter feature space $\mathcal{X}^{(r)}$, $\mathcal{L}^{(r)}$ is the reconstruction loss associated with the auto-encoding approach, 
${\bf{g}}^{(r)}_\phi$ in this task is the decoder and ${\bf{F}}^{(r)}_\theta \equiv {\bf{f}}_\theta$ in the context of the RFF framework is viewed as the projection of the data into the emitter's RF fingerprint representation space. 

In practice, this optimisation problem is solved by collecting an empirical dataset that is assumed to be sampled from $\mathcal{P}_{\mathcal{X}^{(r)}}$ IID $n_r$ times, resulting in a dataset $\mathcal{D}^{(r)} = \{({\bf{x}}^{(r)}_k)\}_{k=1}^{n_{r}}$, where ${\bf{x}}^{(r)}_k \in {\mathcal{\bf{X}}}^{(r)}$ with ${\mathcal{\bf{X}}}^{(r)}$ being the sample dataset. 
This leads to the following optimisation problem is:
\begin{eqnarray}
\xi^*_{em} = \mbox{arg}\min_{\xi\in\Xi}\mathbb{E}_{ {\bf{x}} \sim{\mathcal{D}^{(r)}}}\mathcal{L}^{(r)}\left(  \hat{\bf{x}} , {\bf{x}}\right) \nonumber \\ 
           = \mbox{arg}\min_{\xi \in \Xi} \frac{1}{n_r} \sum_{k=1}^{n_{r}} \mathcal{L}^{(r)}\left(   \hat{\bf{x}}_k, {\bf{x}}_k\right), \nonumber 
\end{eqnarray}
where $\hat{ {\bf{x}}}_k = {\bf{g}}^{(r)}_{{\phi}^{*}} \left({\bf{F}}^{(r)}_{\theta^{*}} \left( {\bf{x}}_k \right) \right)$. To group the processed RF emissions, any clustering algorithm can be applied to the learnt signal representation (i.e. fingerprint as in Figure \ref{mlpipeline}) in a lower dimensional space, this is ${\hat{\bf{z}}}_k = {\bf{F}}^{(r)}_{\theta^{*}} \left( {\bf{x}}_m \right)$ for a given input $\bf{x}_m$,   based on the encoder output; see Section  \ref{sec:RFEC_DMR} for an illustration of RFEC on real RF captures.
\subsection{Open Set SEI and EDA}\label{sec:singletask_OS}
When solving any of the above RFF-dependent tasks,  the generalisation of the ML model to emitter data not contained in the training dataset could be a fundamental challenge. Open Set Recognition (OSR) methods typically first determine whether a sample belongs (or is sufficiently close) 
to the original training data distribution or not. In the CS-SEI regime, a finite number of emitters are assumed in existence and a model is trained to discriminate between them; predicated on the closed world assumption (CWA) \cite{forssell2020equivalence}. 
From the OSR lens, the key difficulty following training in the CS-SEI regime is to establish whether an individual sample presented to the trained model is an observation from a known emitter or an unknown one. The CWA ensures false positives occur on open set data since the model is designed to predict an emitter class from the set of known emitters. In contrast, the critical open set (OS) obstacle in EDA is understanding when the comparator has observed enough emitters in training such that it generalises well to unknown ones. 
In other words, EDA learns to compare and is not conditioned on the emitter class. ~Without loss of generality, we introduce the OSR-SEI problem within the framework in Sections \ref{sec:framework} and \ref{sec:singletasks} using terminology from \cite{vaze2021open}.

Let us assume two sample datasets $({\bf{X}}_{known}^{(i)}, {\bf{Y}}_{known}^{(i)} )$ and $({\bf{X}}_{unknown}^{(i)}, {\bf{Y}}_{unknown}^{(i)})$ are drawn from the joint probability distribution  $\mathcal{P}_{\mathcal{X}^{(i)}\mathcal{Y}^{(i)}}$
such that ${\bf{Y}}^{(i)}_{known} \cap {\bf{Y}}^{(i)}_{unknown} = \emptyset$. ${\bf{Y}}^{(i)}_{known}$ and ${\bf{Y}}^{(i)}_{unknown}$ denote the label set of known and unknown emitters to the CS-SEI model, respectively. 
Then, we define the CS-SEI dataset $\mathcal{D}_c^{(i)} = \left\{ ({\bf{x}}_k^{(i)}, y_k^{(i)}) \right\}_{k=1}^{n_i}$, where ${\bf{x}}_k^{(i)}\in {\bf{X}}^{(i)}_{known}$, $y_k \in {\bf{Y}}^{(i)}_{known}$, 
and $n_i$ is the total number of samples. In addition, we define a OSR-SEI dataset $\mathcal{D}_{o}^{(i)} = \{({\bf{x}}_k^{(i)}, y^{(i)}_k)\}_{k=1}^{n_o} \subset ({\bf{X}}_{known}^{(i)} \bigcup {\bf{X}}_{unknown}^{(i)}) \times ( {\bf{Y}}^{(i)}_{known} \bigcup {\bf{Y}}^{(i)}_{unknown})$ 
which may or may not contain unknown unique emitters, with $n_o$ being the total number of samples. We can combine both into a single dataset as per $\mathcal{D}^{(osr, i)} = \{\mathcal{D}_c^{(i)}, \mathcal{D}_{o}^{(i)}\}$.

In the OSR setting, the goal is to simultaneously learn a model capable of solving the CS-SEI problem defined in Section \ref{sec:singletask_CSSEI} and produce a score indicating whether a sample belongs to $\it{any}$ of the known classes. We define a score function $S(x) \in \{0, 1\}$, which indicates whether a sample $x$
belongs to the class of known emitters or the class of unknown emitters ($0$ indicates $x$ belonging to a known class and 1 otherwise). 
The two most common approaches to determining this score are:
\begin{enumerate}
\item Jointly learn $S(x)$ by using $\mathcal{D}_{c}^{(i)}$ for CS-SEI model training and $\mathcal{D}^{(osr, i)}= \{ \mathcal{D}_{c}^{(i)}, \tilde{\mathcal{D}}_{o}^{(i)} \}$ for binary 
classification between known and unknown emitters, where $\tilde{\mathcal{D}}_{o}^{(i)}$ is a proportion of $\mathcal{D}_{o}^{(i)}$, then evaluate performance on remaining OSR data $\mathcal{D}_o^{(i)} \backslash \tilde{\mathcal{D}}_o^{(i)}$, and;
\item Conduct CS-SEI model training using $\mathcal{D}_{c}^{(i)}$ followed by applying rule-based strategies to compute 
$S(x)$, for example via maximum softmax probability (MSP) with threshold \cite{vaze2021open} \cite{scheirer2012toward} and then employing $\mathcal{D}_{o}^{(i)}$ to assess the OSR performance. 
\end{enumerate}

\begin{figure}[!t]
\centering 
\includegraphics*[width=8cm, height=3.25cm]{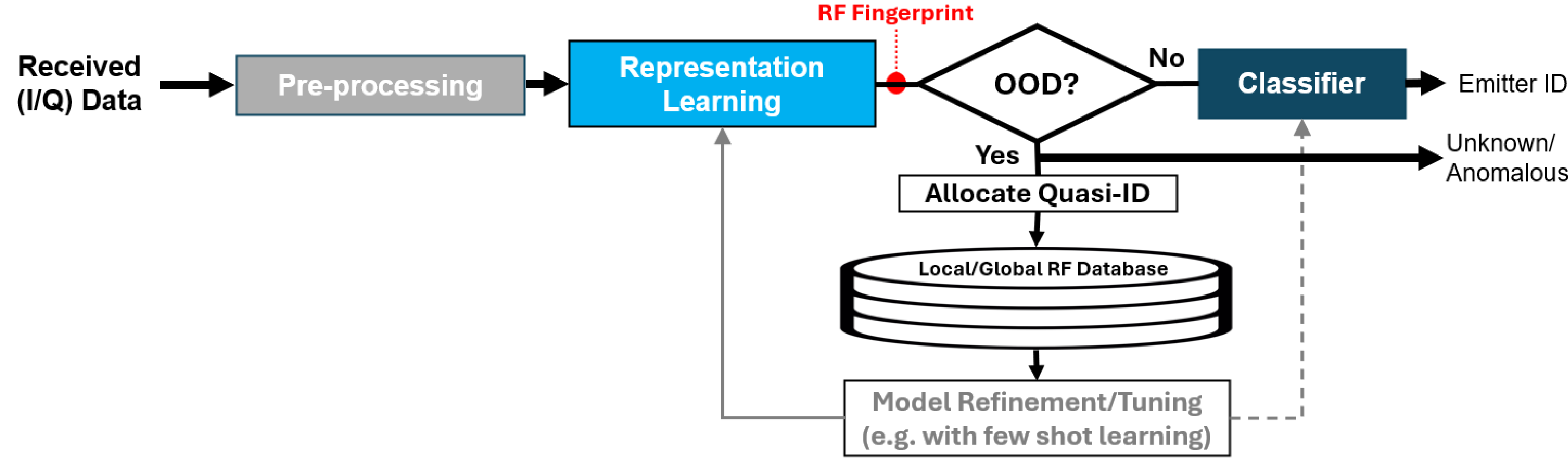}
\caption{Example pipeline for handling OS-SEI.}
\label{example_pipeline}
\end{figure}

Whilst a hybrid of the two techniques can be utilised (e.g. ensemble methods), within the presented generic RFF framework either correspond to solving the following optimisation problem:
\begin{eqnarray}
\min_{\xi \in \Xi} \mathbb{E}_{( {\bf{x}}, {\bf{y}})\sim \mathcal{D}^{(osr, i)}}\mathcal{L}^{(osr, i)}\left( {\bf{h}}_\xi^{(osr, i)}\left({\bf{x}}\right) , y\right), \nonumber
\end{eqnarray}
where ${\bf{h}}^{(osr, i)}_\xi({\bf{x}}) = {\bf{g}}^{(osr, i)}_\phi \left( {\bf{F}}^{(osr, i)}_{\theta}\left({\bf{x}}\right) \right)$, ${\bf{x}} = ( {\bf{x}}_c, {\bf{x}}_o)$ and ${\bf{y}}= (y_c, y_o)$ with $({\bf{x}}_c, y_c)\in \mathcal{D}_{c}^{(i)}$ and $({\bf{x}}_o, y_o) \in \mathcal{D}_{o}^{(i)}$. In this case, ${\bf{g}}_\phi^{(osr, i)}$ is not only the classifier head but also the 
open set score computer and $\mathcal{L}^{(osr, i)}$ could be an OSR-specific objective function such as that in \cite{chen2021adversarial}. Once a methodology for scoring has been established and solved, it can then be used to 
conduct incremental learning \cite{peng2022few} \cite{li2023class}, perform CS-SEI with quasi-labeling of OSR samples \cite{li2025fscil}, using outlier exposure with label smoothing \cite{wang2025sei}, amongst other methods, to address the OS problem. An alternative is meta-learning type techniques which can be leveraged for few-shot open-set recognition as in \cite{xie2025novel}. Figure \ref{example_pipeline} depicts an example pipeline for incorporating OS-SEI. It is underpinned by: a) a suitable out-of-distribution detector (OOD) \cite{yang2024generalized} to reveal out of distribution samples (i.e. those that belong to a emitter not currently in the closed-set list of classes); and b) retraining, including incrementally, the CS-SEI model to address the detected sample(s) from new emitters and their corresponding (quasi-)labels, where relevant. 
\section{Experimentation: Datasets, Models and Set-up} \label{sec:experimentation}
In this section, we first describe the real RF datasets, the ML model architectures and the general experimentation set for a number of disparate RFF-dependent tasks.   
\subsection{Datasets} \label{sec:data}
\subsubsection{Digital Mobile Radios: SIGINT} \label{sec:data_DMR}
The DMR dataset was collected from eight commercial off-the-shelf (COTS) handsets using the COTS software defined radio (SDR) LimeSDR in four environments; 
an open field with direct line-of-sight between each DMR and the SDR, two semi-urban environments and one urban environment. The DMR devices are from five different manufacturers, namely one Hytera PD705, one TYT, 
two Kenwood NX1300D (i.e. identical models), two Entel DX482, and two Motorola DP4400E. All DMRs were programmed to operate in the 433 MHz band where they can use one of 14 channels, each with a bandwidth of approximately 12.5 kHz. 
The sampling rate at the receiver was set at approximately 200 kHz. As part of the data pre-processing steps, a burst segmentation algorithm was applied for segments of 30 seconds of continuous transmission to isolate individual bursts from each DMR handset. 
RFF and identification of individual DMRs has several applications in SIGINT and ISR contexts such as determining the presence of emitters of interest (e.g. associated with high value targets or threats), tracking them over time and/or building pattern-of-life 
of their operators.  
\subsubsection{AIS: Spaceborne Maritime Surveillance}\label{data_AIS}
The automatic identification system dataset contains transmissions from vessels in and near the Panama Canal area as captured by SDRs on Low Earth Orbit (LEO) satellites. 
This proprietary dataset comprises of approximately 10 minutes of RF recordings of the VHF channels dedicated to ground maritime AIS emitters.  
Bespoke pre-processing algorithms are applied to detect-extract the individual AIS bursts and, where possible, demodulate-decode them to obtain their unique maritime mobile service identity (MMSI) number, which can serve as a label. 
It is noted that classical demodulating-decoding techniques for AIS often require relatively high-quality AIS transmissions, for instance a substantially high SNR and/or low levels of interference. 
The AIS dataset utilised here has over 600 AIS bursts from 48 sources such that each emitter has at least 10 detected transmissions. The sampling rate is approximately 38.4 kHz with an AIS burst time slot of 26.67ms. RFF-based EDA can improve geolocating the transmission source. By identifying various transmissions that originate from the same source, we can effectively constrain the observables (time and frequency difference of arrival measurements) and dwell on the emitter over a much longer time intervals. Thereby, we can utilise multiple emissions in the estimation of the emitters’ locations. Whilst geolocation of ground RF emitters from LEO satellite clusters and subsequently tracking changes in their positions  (including from multiple passes) has many ISR applications, spaceborne RFF facilitates delivering enhanced situational awareness. For example, results from SEI or EDA can be fused with other data sources, such as images, to confirm the identity of vessels and/or detect malicious activities (e.g. AIS spoofing).
\subsubsection{RF Data from Drones: Counter-UAS} \label{sec:data_drones}
RF captures from four identical (i.e. same manufacturer, model and year) small hobbyist drones, Chubory F89, were collected with a COTS SDR in two different settings. These drones use the 2.4 GHz WiFi spectrum for the uplink (controller to drone) and downlink (drone to controller such as video feed) with frequency hopping within. The uplink and downlink transmissions had a central frequency of 2.47GHz, processed bandwidth is approximately 28 MHz to cover several of the subbands (namely, on the uplink transmission) and the SDR sampling rate was set at 56MHz. The dataset contains approximately 40,000 emissions per drone.
Our objective is to demonstrate that ML-based RFF enables the identification of each individual drone controller from raw I/Q data, in scenarios where the transceivers are the same make and model. This goes well beyond broader classification tasks that might aim to only determine the type of the drone (e.g. based on traditional signal or modulation-scheme classification) or drone versus non-drone RF emissions. 
In this context, SEI can facilitate threat detection and pattern-of-life analysis of drone(s) operators in a C-UAS context. These capabilities can offer a substantial operational advantage in various scenarios, inclusive of cueing C-UAS effectors and resource planning.    
\subsection{Model Architectures}\label{sec:models}
The experimentation conducted in this paper uses example common ML model architectures that compose of convolution \cite{lecun2002gradient}, long short term memory (LSTM) \cite{hochreiter1997long}, transformer \cite{vaswani2017attention} and fully connected operations. They are:  
\begin{enumerate}
\item {\bf{Block convolutional neural network (BCNN)}}: multiple blocks of one-dimensional CNNs, batch normalisation \cite{ioffe2015batch}, leaky rectified linear unit \cite{maas2013rectifier} and max pooling layers. Five convolutional blocks are followed by two linear layers with an optional softmax layer; 
\item {\bf{Bidirectional LSTM (BILSTM)}}: a LSTM-based architecture \cite{graves2013speech}. Specifically, we use the PyTorch implementation with the following parameters: $\mbox{input\_size}=2, \mbox{hidden\_size}=512, \mbox{num\_layers}=2, \mbox{bidirectional}=\mbox{True}$, with a linear layer attached at the end of the model;
\item {\bf{GLFormer}}: a transformer-based architecture for CS-SEI introduced in \cite{deng2023lightweight}. The architecture employs the following mechanisms: sequence embedding, gated attention units (GAUs), local self-attention (LSA), sliding local self-attention (SLA) \cite{pan2023slide} and global average pooling (GAP) \cite{lin2013network};
\item {\bf{VGG19}}: one-dimensional variant of VGG19 \cite{simonyan2014very};
\item {\bf{Fully connected network (FCN)}}: sequence of linear layers and leaky rectified linear unit activation functions.
\end{enumerate}
For RFEC, we used four different auto-encoder architectures which we named: simpleAE, verysimpleAE, simpleconv1DAE and vanillaAE. 
The former model is comprised of a four layer fully connected network across both the encoder and decoder with leaky ReLU activation after each layer. 
The verysimpleAE model is the same as simpleAE, however with only one layer. The simpleconv1DAE parameterises the encoder with three one-dimensional convolutional layers with batch normalisation and leaky ReLU activation followed by three layers of fully connected networks in the encoder and decoder. 
The vanillaAE architecture is another FCN, but with three layers in the encoder and decoder, batch normalisation and leaky ReLU activation. The standard K-means clustering algorithm is applied to the learnt low-dimensional representations (i.e. outputs of the encoder).

It is emphasised that other suitable ML model architectures can be applied within the presented generic RFF framework for any of the considered tasks. 

\subsection{Setup} \label{sec:setup}
For all three datasets described in Section 5.1, and for all three tasks (SEI, EDA and RFEC), the input to the ML models is the received transmission raw I/Q data. We split each dataset into training, validation and testing data. In particular, 
$\mathcal{D}^{(t)}=\{\mathcal{D}^{(t)}_{train}, \mathcal{D}^{(t)}_{valid}, \mathcal{D}^{(t)}_{test}\}$ where $|\mathcal{D}^{(t)}_{train}| = 1- p_v - p_t$, $|\mathcal{D}^{(t)}_{valid}|=p_v$ and $|\mathcal{D}^{(t)}_{test}|=p_t$ such we choose here $p_v=p_t=0.1$. Across all datasets, we use the multi-class cross-entropy loss function for SEI, the contrastive loss function for EDA and the mean-squared error loss for RFEC. For a single data sample, these are summarised as:
\begin{eqnarray}
\mathcal{L}^{(i)}({\bf{\hat{y}}}, {\bf{y}}) = -\sum_{j=1}^{\mathcal{V}} y_j \log(\hat{y}_j); \nonumber \\
\mathcal{L}^{(a)}(\hat{y}, y) = (1-y) \lvert \lvert {\bf{f}}_\theta (x_1) - {\bf{f}}_\theta(x_2) \rvert \rvert _2^2 \nonumber \\
                              + y \max (0, m- \lvert \lvert {\bf{f}}_\theta (x_1) - {\bf{f}}_\theta (x_2)\rvert \rvert _2^2); \nonumber \\
\mathcal{L}^{(r)}({\bf{\hat{y}}}, {\bf{y}})    =  \lvert \lvert {\bf{\hat{y}}} - {\bf{y}} \rvert \rvert _2^2, \nonumber 
\end{eqnarray}
where $y_j$ is the ground-truth probability for emitter class $j$, $\hat{y}_j$ the predicted probability for the $j$th emitter class, $y$ and $\hat{y}$ in $\mathcal{L}^{(a)}$ are the ground-truth match and predicted match respectively, 
${\bf{y}}$ and ${\bf{\hat{y}}}$ are the ground-truth data and encoding of the data vectors in $\mathcal{L}^{(r)}$, $m$ is the margin parameter in the contrastive loss function, and $\lvert \lvert \cdot \rvert \rvert _2$ is the $L_2$ norm. In order to update the parameters which minimise these loss functions in practice, across the appropriate datasets, we define the iterative parameter update formula:
\begin{equation}
\xi_t^k = \mathcal{U}\left(\tau^{(t)}_k, \mathcal{L}^{(t)}\left( {\bf{\hat{Y}}}^{(t)}_{k}, {\bf{Y}}^{(t)}\right) \right) \mbox{for} \; k = 1, 2, \cdots, K, \nonumber 
\end{equation}
where $\xi_{t}^k = (\phi_t^k, \theta_t^k)$, with $k$ referring to the current iteration, $t$ the task, $\tau^{(t)}_k$ the learning rate, ${\bf{\hat{Y}}}_{k}^{(t)}$ the model prediction, ${\bf{Y}}^{(t)}$ the ground-truth label set
and $\mathcal{L}^{(t)}$ the loss function. Here, we adopt mini-batch (MB) gradient descent to update the model parameters. 
Algorithm \ref{MB_STL} shows pseudo-code for MB-single task learning (MB-STL), which is executed for each individual task. 
\begin{algorithm}[!t]
  \begin{algorithmic}
  \State {\bfseries INPUTS:} $\mathcal{T} = (\mathcal{D}, {\bf{g}}_\phi, {\bf{F}}_\theta, \mathcal{L}, \mathcal{U})$
  \State {\bfseries INITIALISE:} $\xi_{0}, \tau_{0}$
  \For{Epoch $e=0$ {\bfseries to} $E-1$}
  \For{batch $b=0$ {\bfseries to} $B-1$}
  \State ${\bf{\hat{y}}}_{e,b} = {\bf{h}}_{\xi_{e, b}}({\bf{X}}_b)$
  \State $\xi_{e+1, b+1} = \mathcal{U}\left(\tau_{e}, \mathcal{L}\left( {\bf{\hat{Y}}}_{e,b}, {\bf{Y}}_{b}\right)\right)$
  \EndFor 
  \EndFor  
  \State {\bfseries OUTPUT:} $\xi_{v, B}^*$ with an early stopping criterion
  \end{algorithmic}
  \caption{MB-STL pseudo-code with stopping criterion}
  \label{MB_STL}
\end{algorithm}
We adopt the Adam optimiser method \cite{kingma2017adammethodstochasticoptimization} as $\mathcal{U}$, set the batch size $b_{size} = 512, 128, 128$ for SEI, EDA, and RFEC, respectively, and assume an epoch $k$ is the completion of 
$ \lceil \lvert \mathcal{D}_{train} \rvert / b \rceil$ batch updates, where $\left \lceil \cdot \right \rceil$ is the ceiling function. We choose the number of total epochs to be $E = 200$ unless specified otherwise. Once the model parameter update is calculated via optimising the loss function, 
it is mapped into a new parameter update that minimises the $L_2$ norm of the parameters $\xi$. This is defined as a weight decay parameter and it is set to $5 \times 10^{-4}$.
The learning rate is constant across training, with $\tau^{(i)}=7.5 \times 10^{-3}, 1 \times 10^{-3}$ (depending on model architecture), 
$\tau^{(a)}= 1 \times 10^{-3}$ and $\tau^{(r)}= 1 \times 10^{-3}$; all chosen empirically. We apply early stopping criteria using the validation dataset $\mathcal{D}_{valid}$, which
retains the model with the best performance across all epochs. It is noted that other set ups and configurations, inclusive of loss function and optimiser, can be employed within the introduced ML framework.

\begin{table}[b]
  \centering
  \tabcolsep=0.11cm
  \begin{tabular}{|l|c|c|c|r|}
  \hline
  \textbf{Architecture} & \textbf{Accuracy} & \textbf{F1 Score} & \textbf{Precision} & \textbf{Recall} \\ \hline
      BCNN        & 91.68\% & 0.915 & 0.922 & 0.917 \\ \hline
      BILSTM      & 88.32\% & 0.881 & 0.906 & 0.884 \\ \hline
      VGG19       & 90.98\% & 0.911 & 0.913 & 0.910 \\ \hline
      GLFormer    & 87.30\% & 0.879 & 0.906 & 0.873 \\ \hline
      FCN         & 89.78\% & 0.898 & 0.899 & 0.898 \\ \hline
  \end{tabular}
  \caption{Accuracy, F1 score, precision and recall of each trained model architecture evaluated on test dataset for SEI (8 handsets)}
    \label{tab:results_cs_sei}
  \end{table}
\begin{figure*}[!htb]
\centering
\includegraphics[width=5.5cm, height=5.5cm]{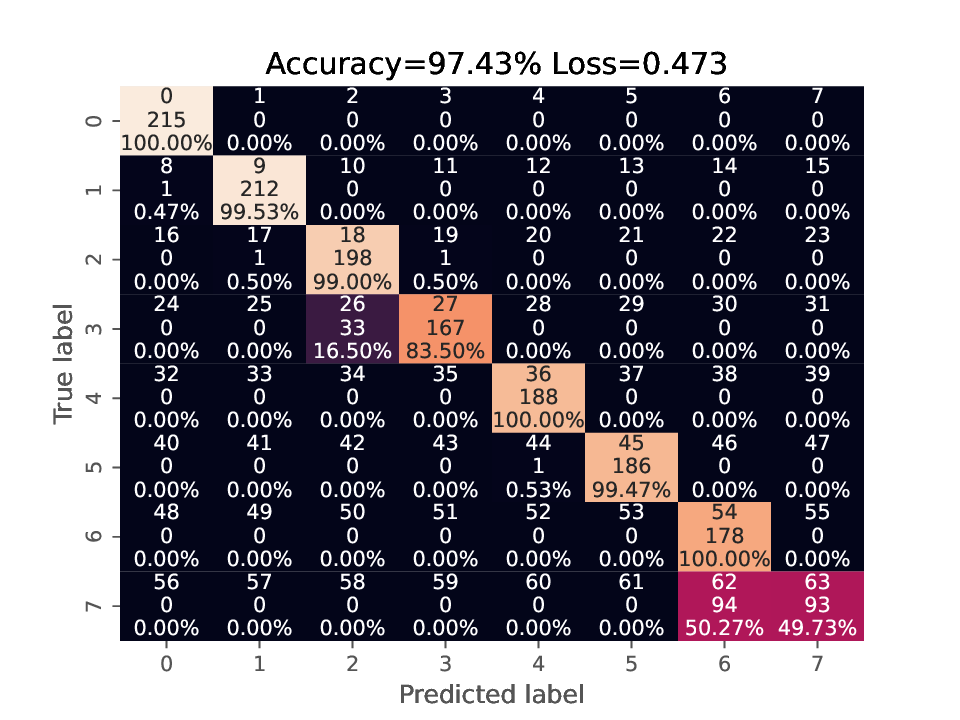}
\includegraphics[width=5.5cm, height=5.5cm]{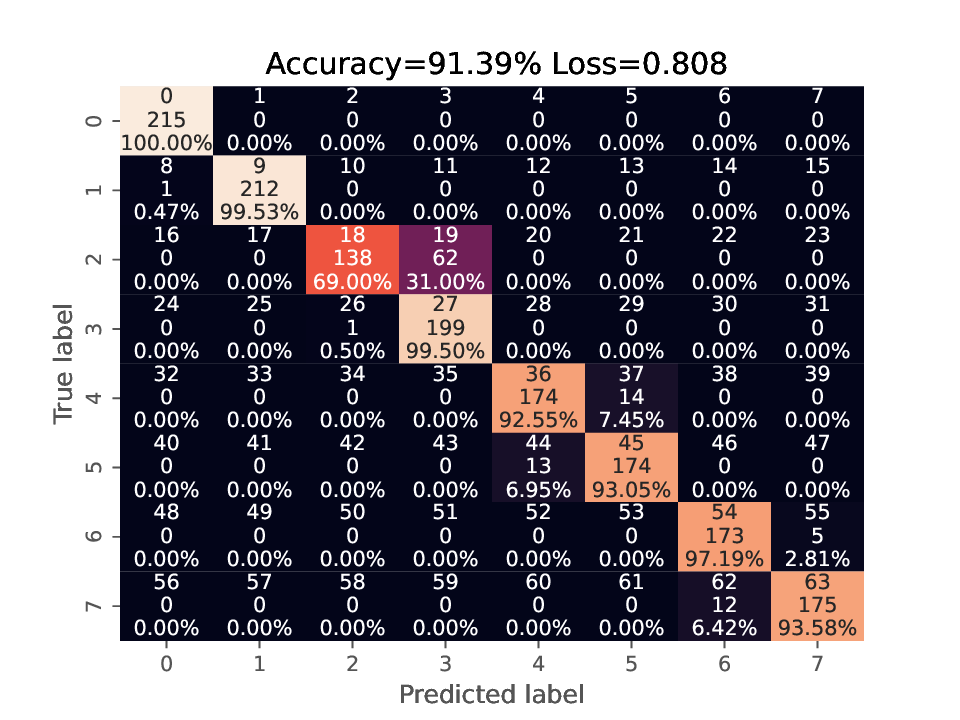}
\includegraphics[width=5.5cm, height=5.5cm]{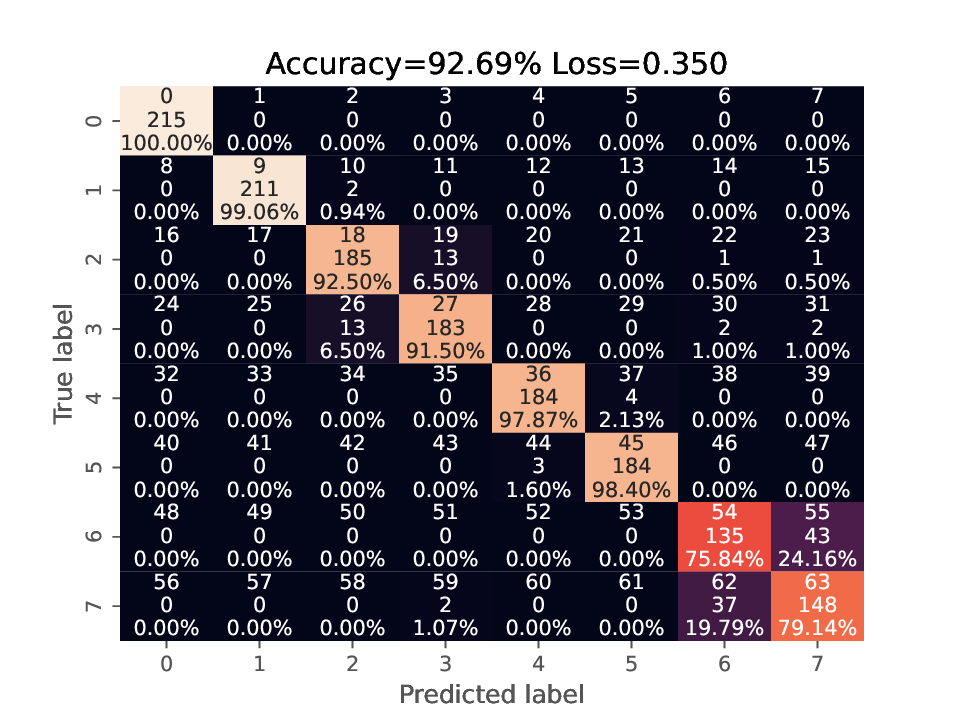}
\includegraphics[width=5.5cm, height=5.5cm]{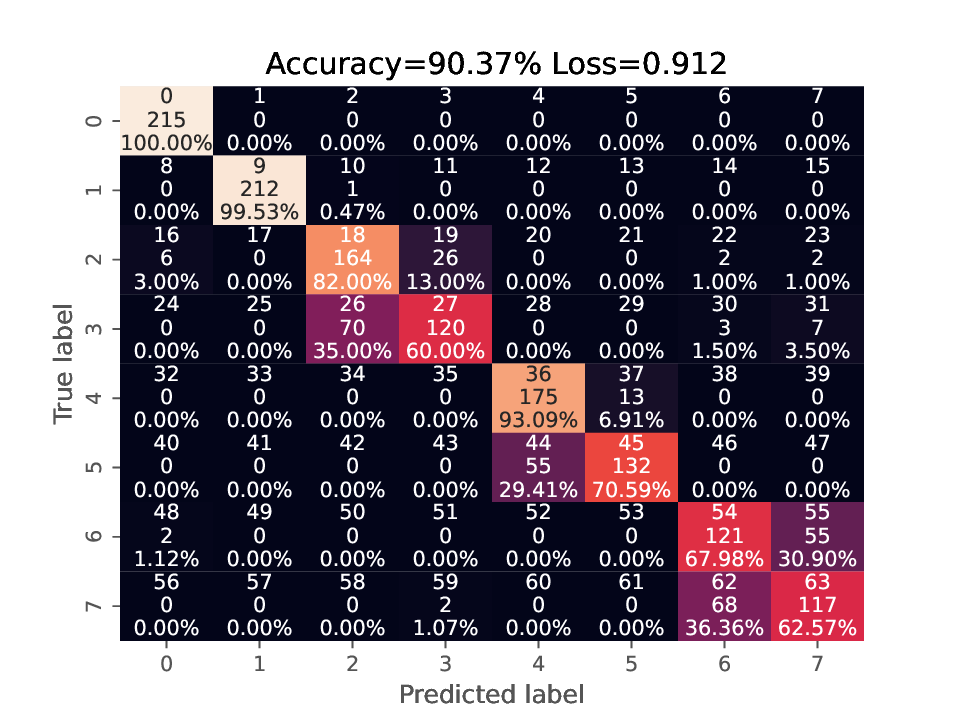}
\includegraphics[width=5.5cm, height=5.5cm]{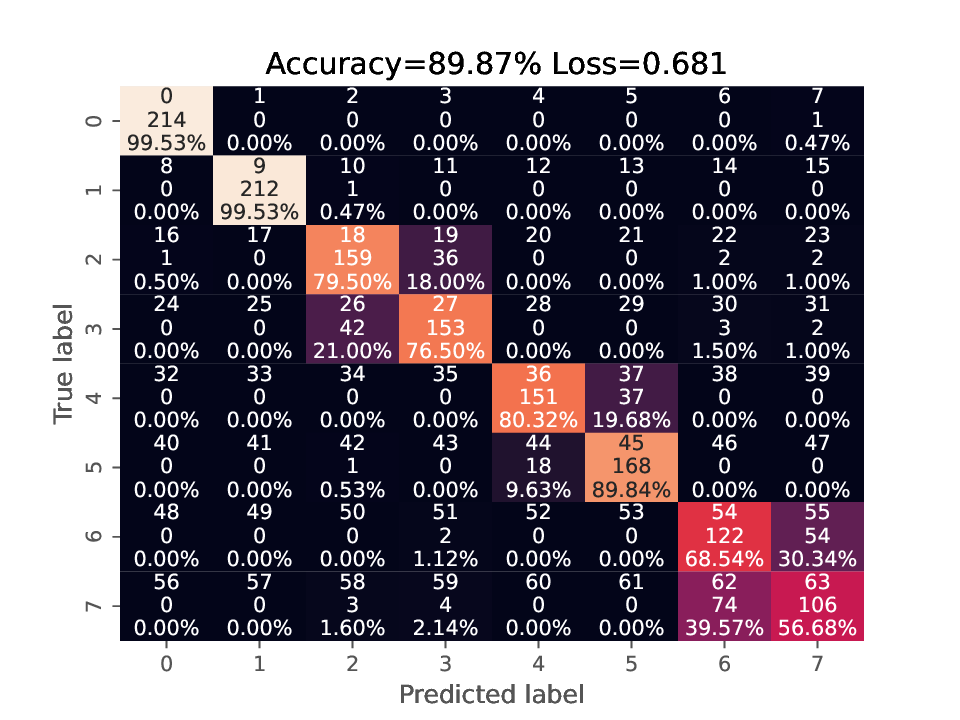}
\caption{CS-SEI test dataset evaluation: confusion matrices. Top: VGG19 (L), BCNN (M), GLFormer (R). Bottom: BILSTM (L), FCN (R)}
\label{fig:bcnn_cm_sei}
\end{figure*}
\begin{figure*}[!htb]
\centering
\includegraphics[width=4.8cm, height=5.5cm]{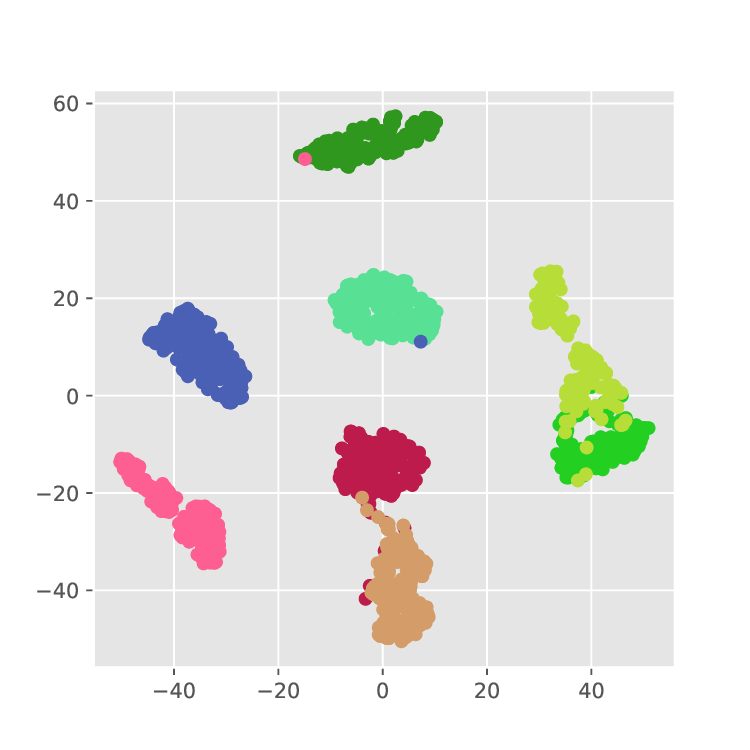}
\includegraphics[width=4.8cm, height=5.5cm]{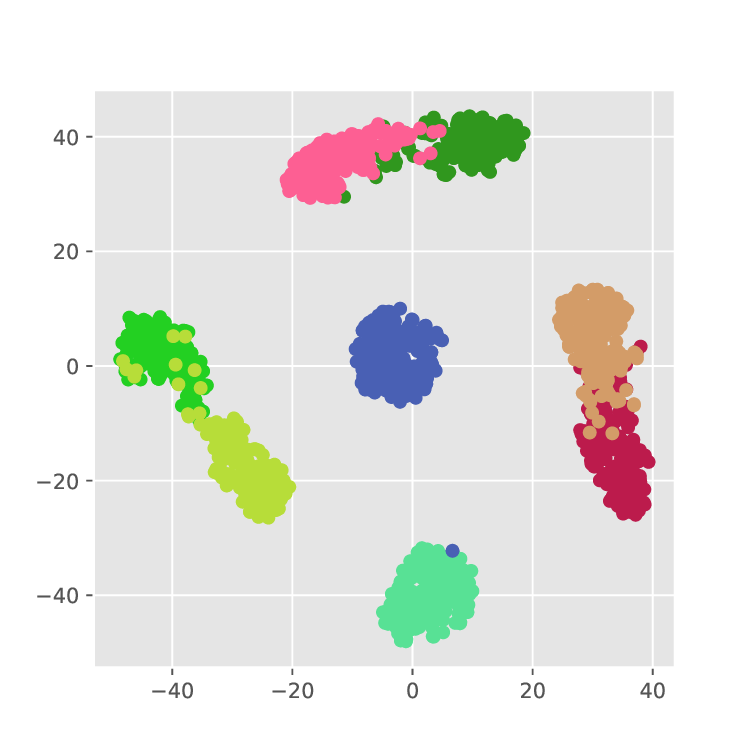}
\includegraphics[width=4.8cm, height=5.5cm]{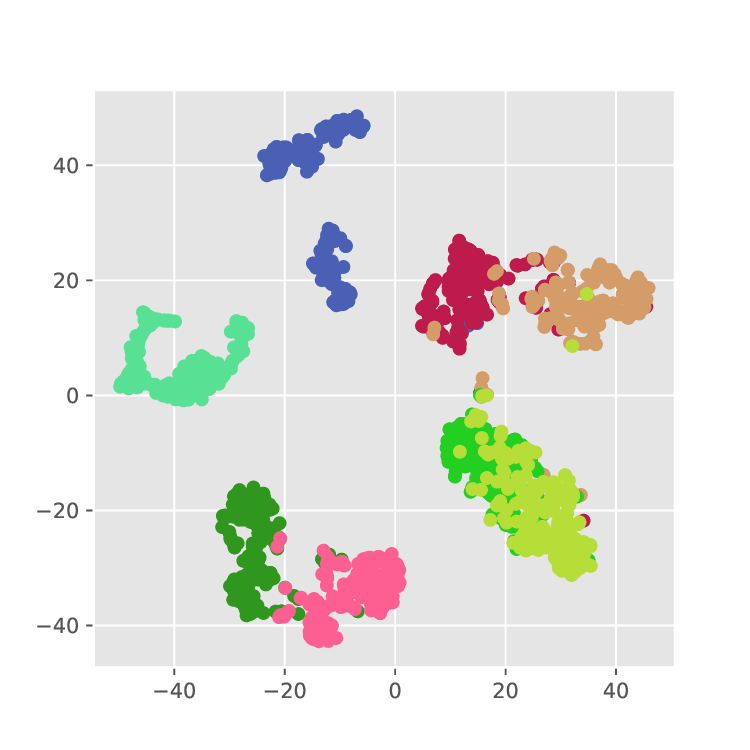}
\includegraphics[width=4.8cm, height=5.5cm]{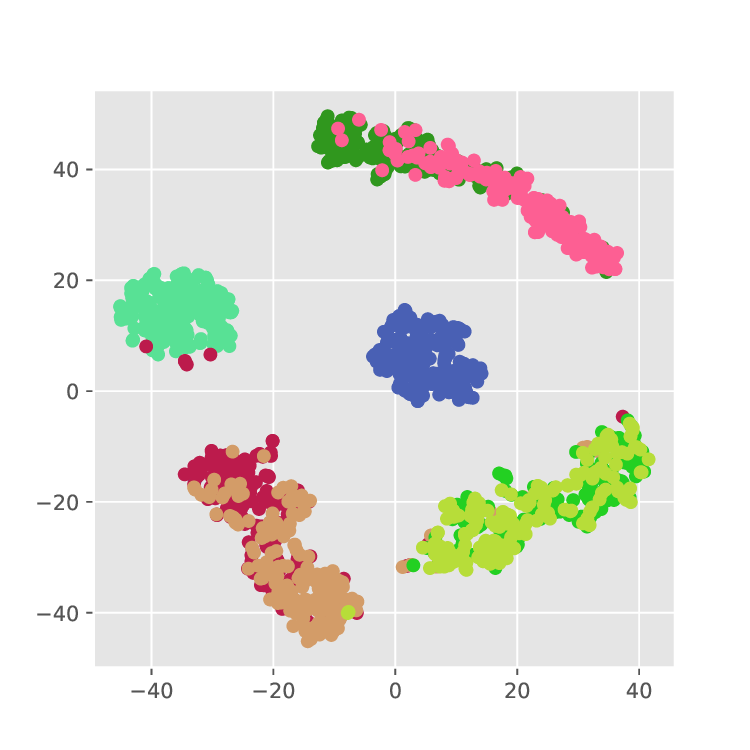}
\includegraphics[width=4.8cm, height=5.5cm]{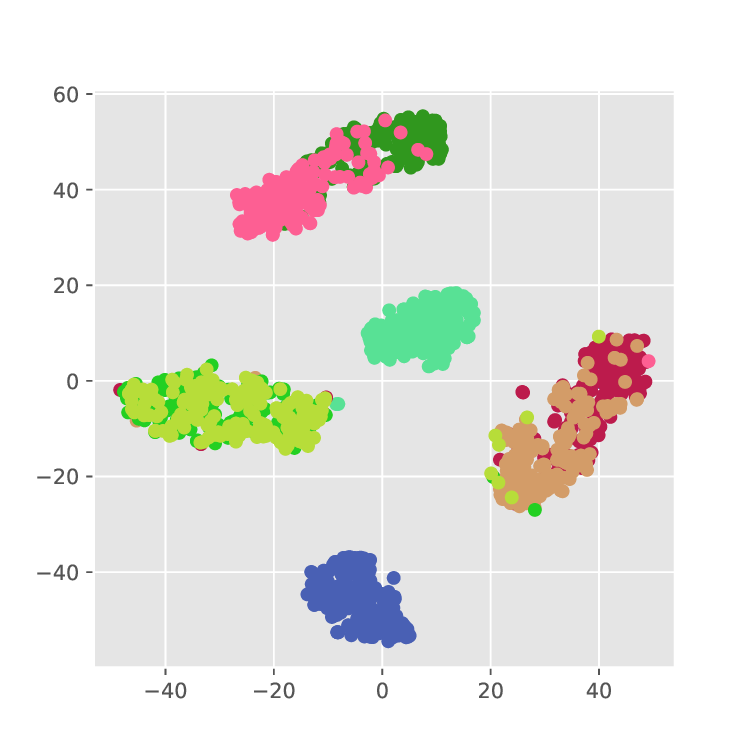}
\caption{CS-SEI Test dataset evaluation: t-SNE data embeddings. Top: VGG19 (L), BCNN (M), GLFormer (R). Bottom: BILSTM (L), FCN (R).}
\label{fig:bcnn_tsne_sei}
\end{figure*}

\section{Results} \label{sec:results}
Results for CS-SEI, EDA, RFEC and OSR-EDA from the datasets in Section \ref{sec:data} are presented here. 
Where applicable, we display confusion matrices and quantitative metrics such as accuracy, precision, recall, F1 Score and Silhouette scores on K-Means clusters. It is noted that the learnt RF fingerprints representations (i.e. embeddings) are typically of dimension 100 or above. For visualisation purposes, we use t-distributed stochastic neighbor embedding (t-SNE) \cite{maaten2008visualizing} for dimensionality reduction to display the fingerprint in 2-D; others such as UMAP \cite{mcinnes2018umap} can be employed.  
\subsection{Use Case 1: CS-SEI for DMR Data} \label{sec:SEI_DMR}
Table \ref{tab:results_cs_sei} provides a summary of CS-SEI results (i.e. metrics accuracy, F1 score, precision and recall) with BCNN, BILSTM, VGG19, GLFormer and FCN model architectures for the DMR test dataset. The corresponding confusion matrices are depicted in Figure \ref{fig:bcnn_cm_sei}. It can be noticed that all of the examined ML models deliver comparable SEI performance, albeit the largest model VGG19 (i.e. in terms of model size and number of parameters) has a marginally higher accuracy compared to a substantially simpler-smaller architectures (e.g. FCN and BCNN). This concurs with results in \cite{bothereau2025rf}. 

Figure \ref{fig:bcnn_tsne_sei} displays the t-SNE plots (i.e. RF fingerprint representation in 2-D). Different colours in each plot refer to a different emitter class and each point within each plot is obtained from a data sample (i.e. time-series of I/Q RF signal from a given transmission for example of length 512).  It can be noted, the RF fingerprint representations from the eight emitter classes are separable across all of the assessed ML models, despite some overlap. The most visible overlaps in fact belong to DMR handsets that are of the same model and make (i.e. from the same manufacturer). Whilst this is expected and implies that there are crossover features for a specific model, their emissions can still be individually identified by their RF fingerprints, especially if this is over several data samples. The overlapping emitter embeddings behaviour from the t-SNE plot can also be observed in Figure \ref{fig:bcnn_cm_sei}, where confusion matrices are displayed for all the trained model architectures on this task. 
Emitters (2,3), (4,5) and (6,7) are unique devices but with the same make and manufacturer, and all architectures have some difficulty in fully separating their class boundaries whereas, for the most part, emitters 0 and 1 can be distinguished with better accuracy. 
  
\subsection{Use Case 2: CS-SEI for Drone RF Data} \label{sec:SEI_Drone}
\begin{figure}[!t]
\centering
\includegraphics[width=4.250cm, height=4.25cm]{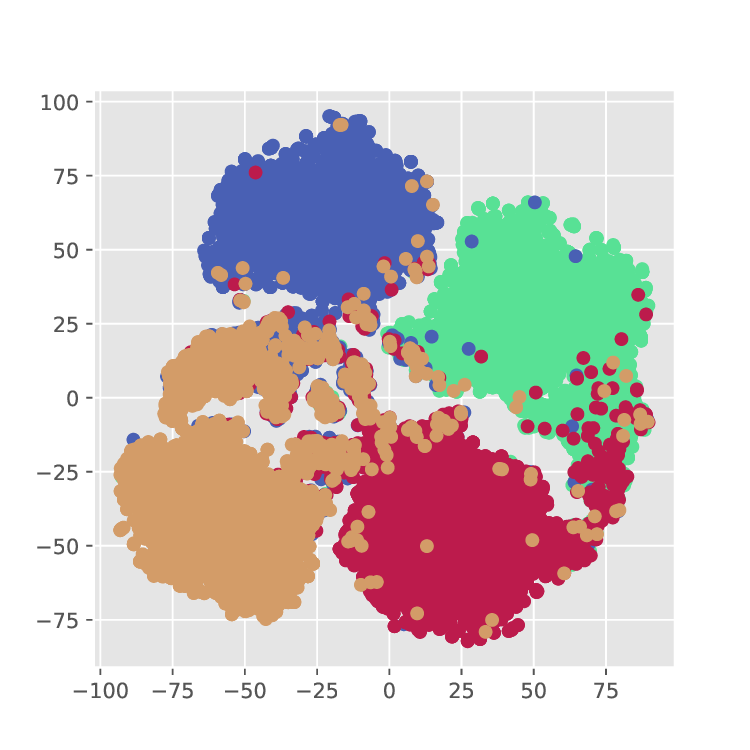}
\includegraphics[width=4.250cm, height=4.25cm]{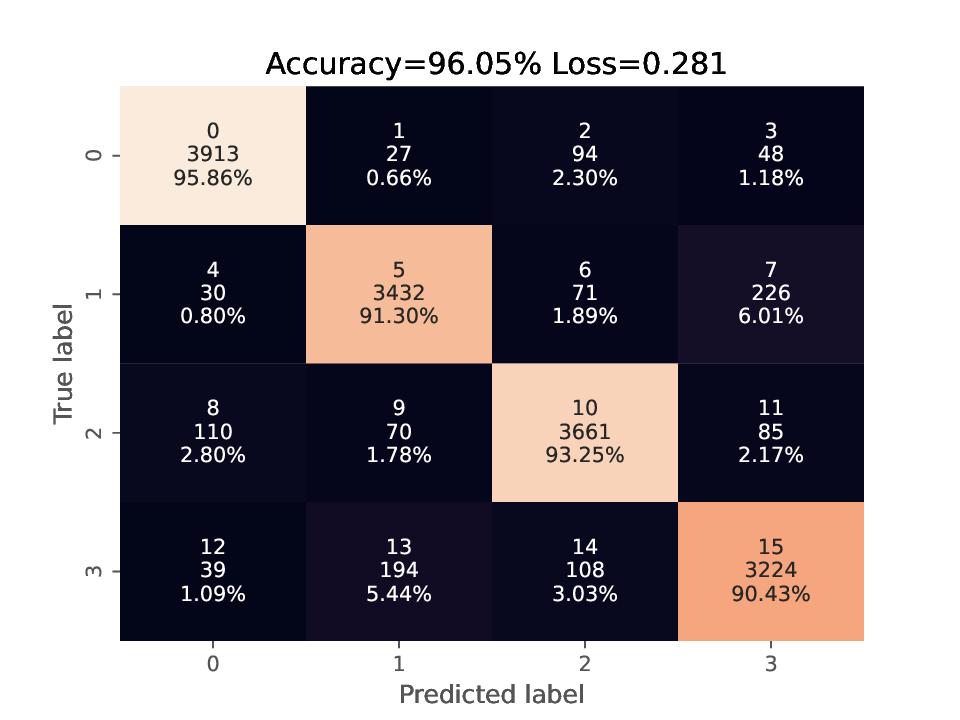}
\caption{CS-SEI for the drone RF links test dataset. Top: t-SNE embeddings using FCN. Bottom: confusion matrix.}
\label{drone_embeddings_cm}
\end{figure}

Figure \ref{drone_embeddings_cm} shows the t-SNE embeddings plot and a confusion matrix of the trained FCN CS-SEI model for the drone RF links dataset in Section \ref{sec:data_drones}. The fingerprints for each of the four drones (i.e. same model and make, thus four different instances of the same transmitter ) are visibly separable with some overlap. The confusion matrix demonstrates consistently good performance across all classes, achieving an overall accuracy exceeding $96\%$ with the basic FLC model. 
\subsection{Use Case 3: EDA for DMR Data}
\begin{figure}[t]
\centering
\includegraphics[width=4.2cm, height=4.2cm]{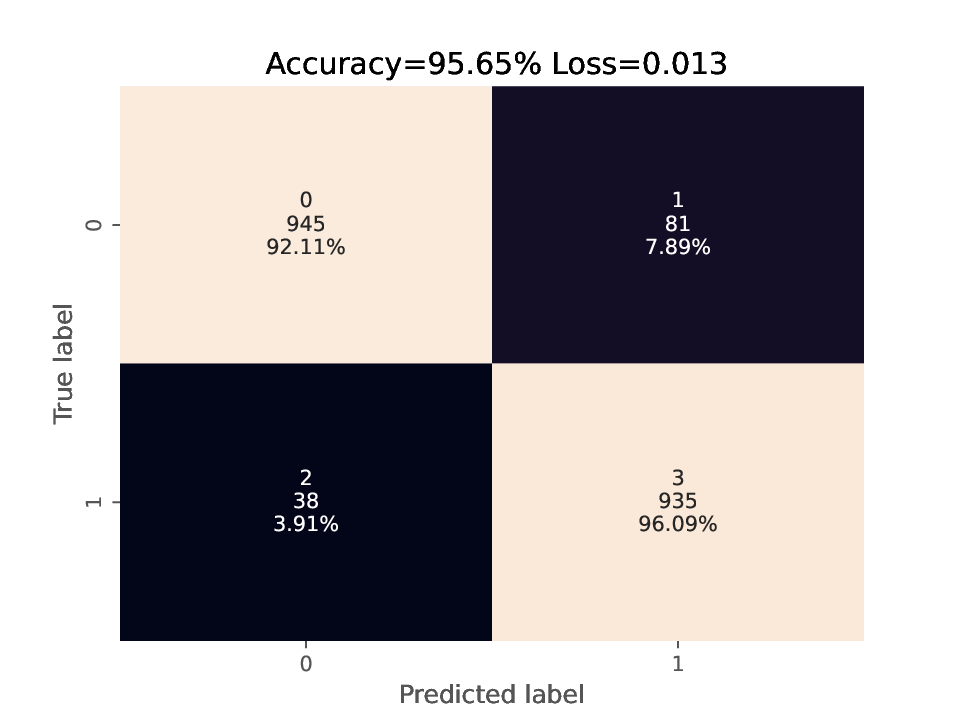}
\includegraphics[width=4.2cm, height=4.2cm]{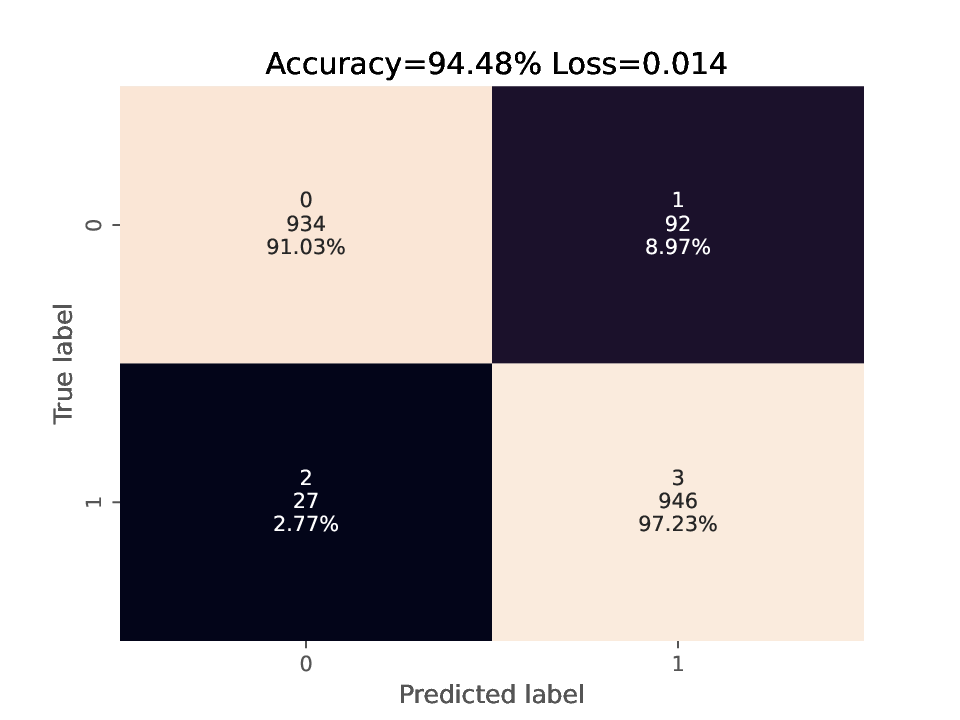}
\includegraphics[width=4.2cm, height=4.2cm]{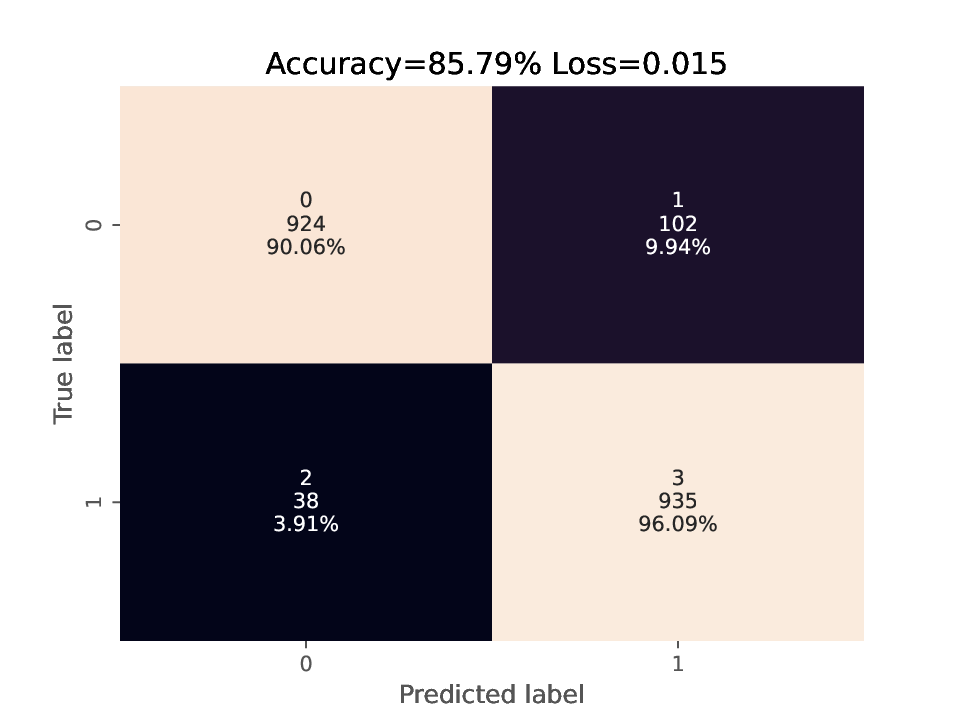}
\includegraphics[width=4.2cm, height=4.2cm]{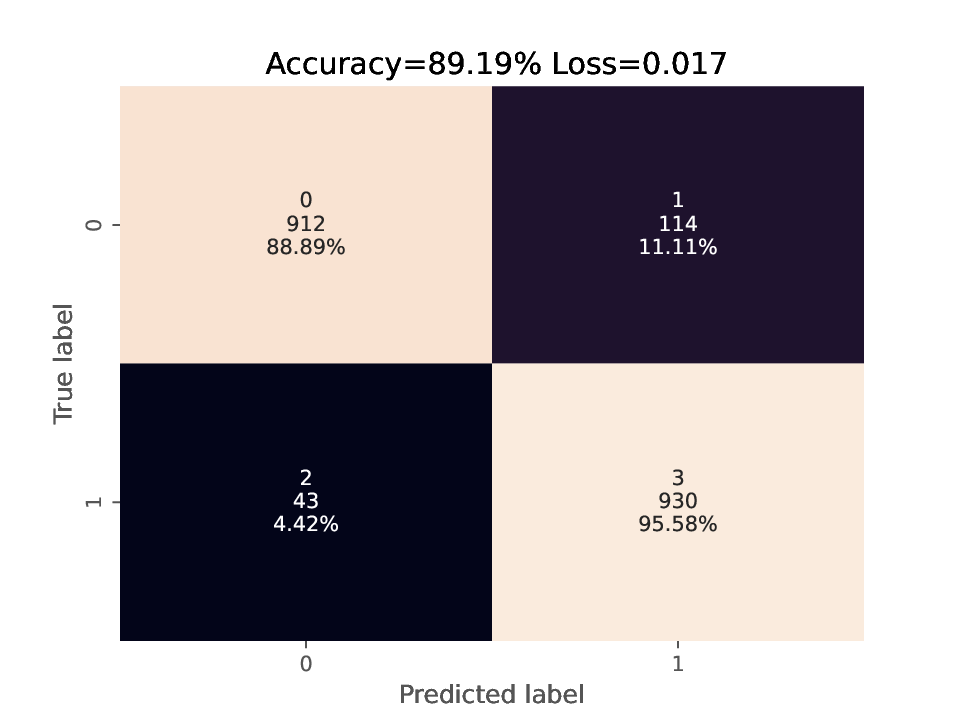}
\caption{EDA test dataset evaluation: confusion matrices. Top: VGG19 (L), BCNN (R). Bottom: GLFormer (L), BILSTM (R).}
\label{bcnn_cm_eda}
\end{figure}

The EDA results for the DMR dataset is depicted in Figure \ref{bcnn_cm_eda}, displaying the four confusion matrices for the VGG19, BCNN, GLFormer and BILSTM models. Table \ref{tab:results_EDA} lists the performance metrics. Both shows good performance across all of the considered architectures. 
\begin{table}[!t]
\centering
\tabcolsep=0.11cm
\begin{tabular}{|l|c|c|c|r|}
\hline
\textbf{Architecture} & \textbf{Accuracy} & \textbf{F1 Score} & \textbf{Precision} & \textbf{Recall} \\ \hline
BCNN & 94.48     \% & 0.940 & 0.910 & 0.972 \\ \hline
BILSTM & 89.19   \% & 0.921 & 0.889 & 0.955 \\ \hline
VGG19 & 95.65    \% & 0.941 & 0.921 & 0.961 \\ \hline
GLFormer & 85.79 \% & 0.923 & 0.901 & 0.960 \\ \hline
\end{tabular}
\caption{EDA performance of trained BCNN, BILSTM, VGG19, GLFormer and FCN models for EDA dataset.}
\label{tab:results_EDA}
\end{table}

\subsection{Use Case 4: EDA for Spaceborne AIS Data}\label{sec:EDA_AIS}
\begin{figure}[!t]
  \centering
  \includegraphics[width=4.25cm, height=4.25cm]{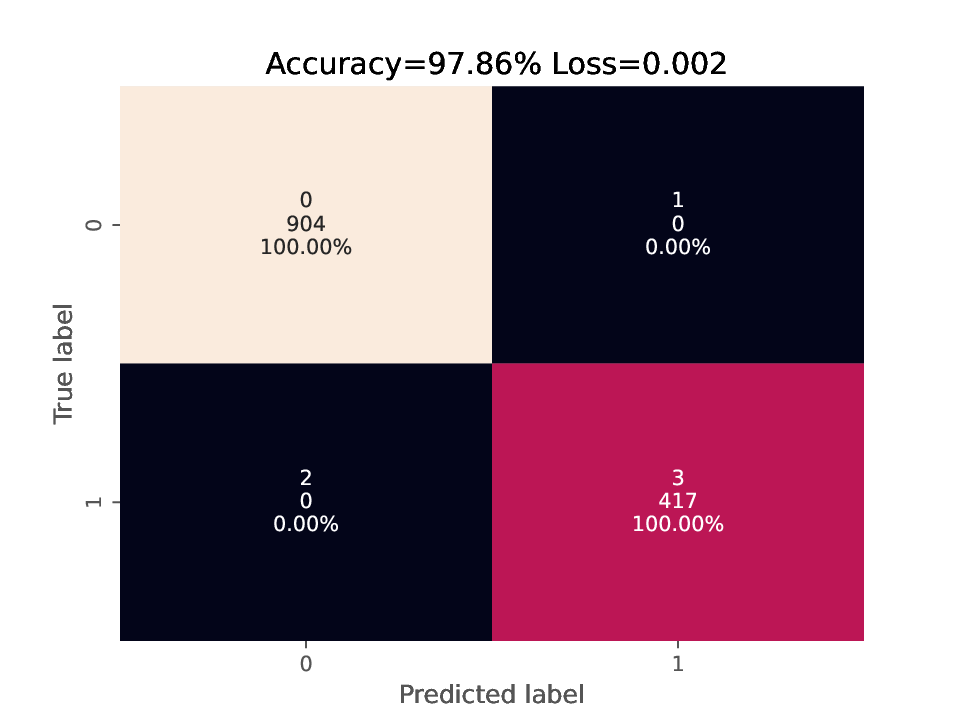}
  \includegraphics[width=4.25cm, height=4.25cm]{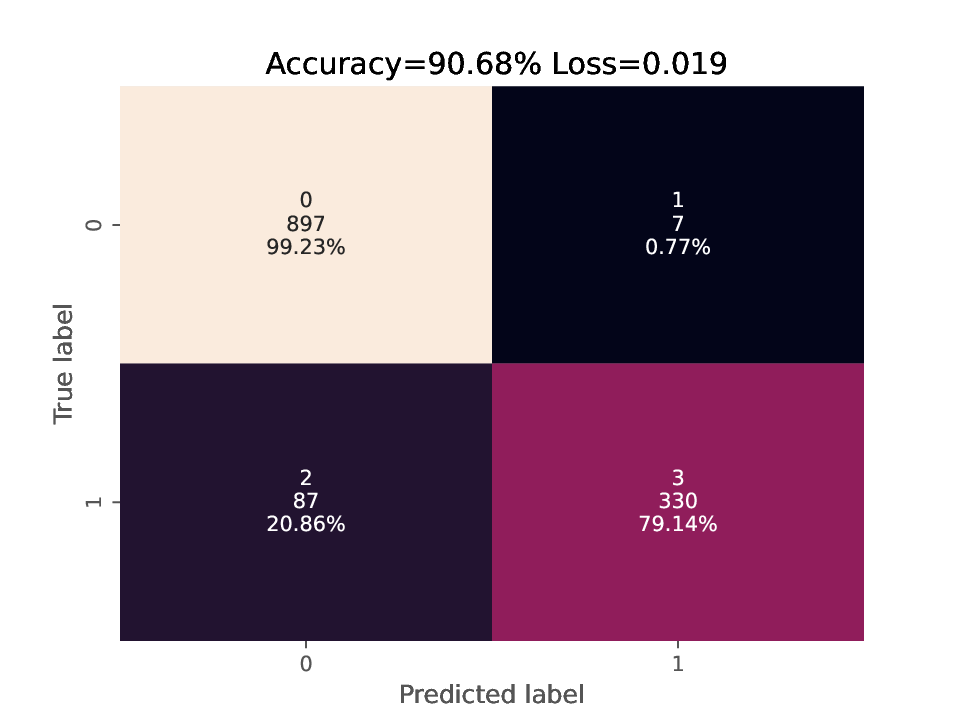}
  \includegraphics[width=4.25cm, height=4.25cm]{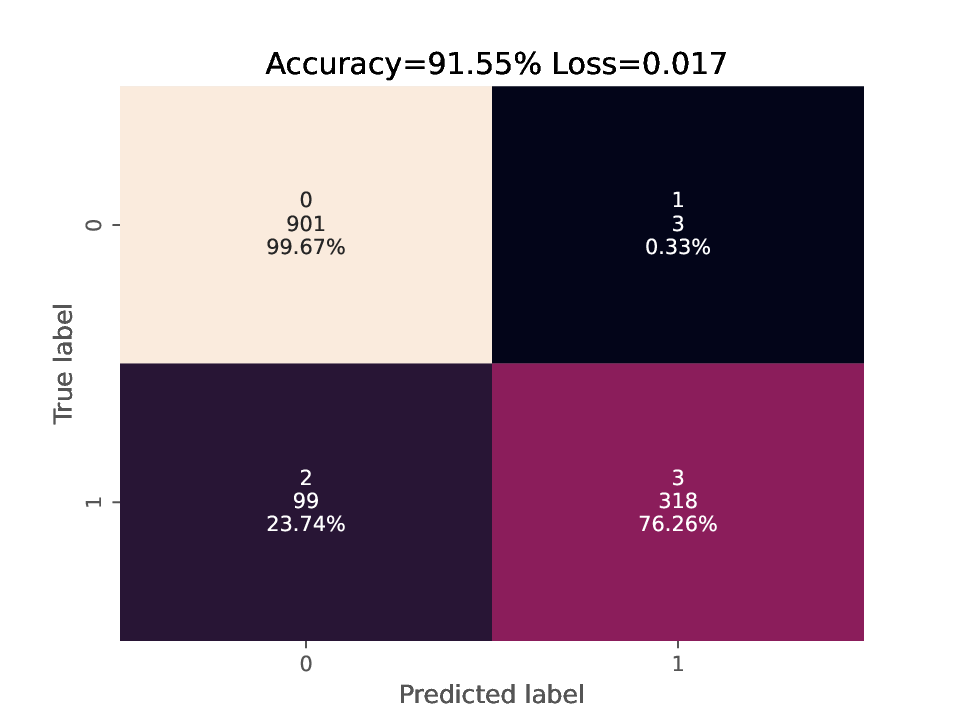}
  \caption{EDA AIS test dataset evaluation: confusion matrices. Top: BCNN (L), BILSTM (R). Bottom: FCN.}
  \label{bcnn_ais_embeddings_eda2}
  \end{figure}

Data from all of the 48 AIS emitters collected by the LEO satellites are used during the training and testing procedure of the three selected models: BCNN, BILSTM and FCN. Figure \ref{bcnn_ais_embeddings_eda2} displays their confusion matrices. Similar to the other experiments above, this figure illustrates that ML-based RFF can deliver a high accuracy EDA for various model architectures, including for spaceborne maritime surveillance. 

\begin{figure}[!t]
  \centering
  \includegraphics[width=5.5cm, height=5cm]{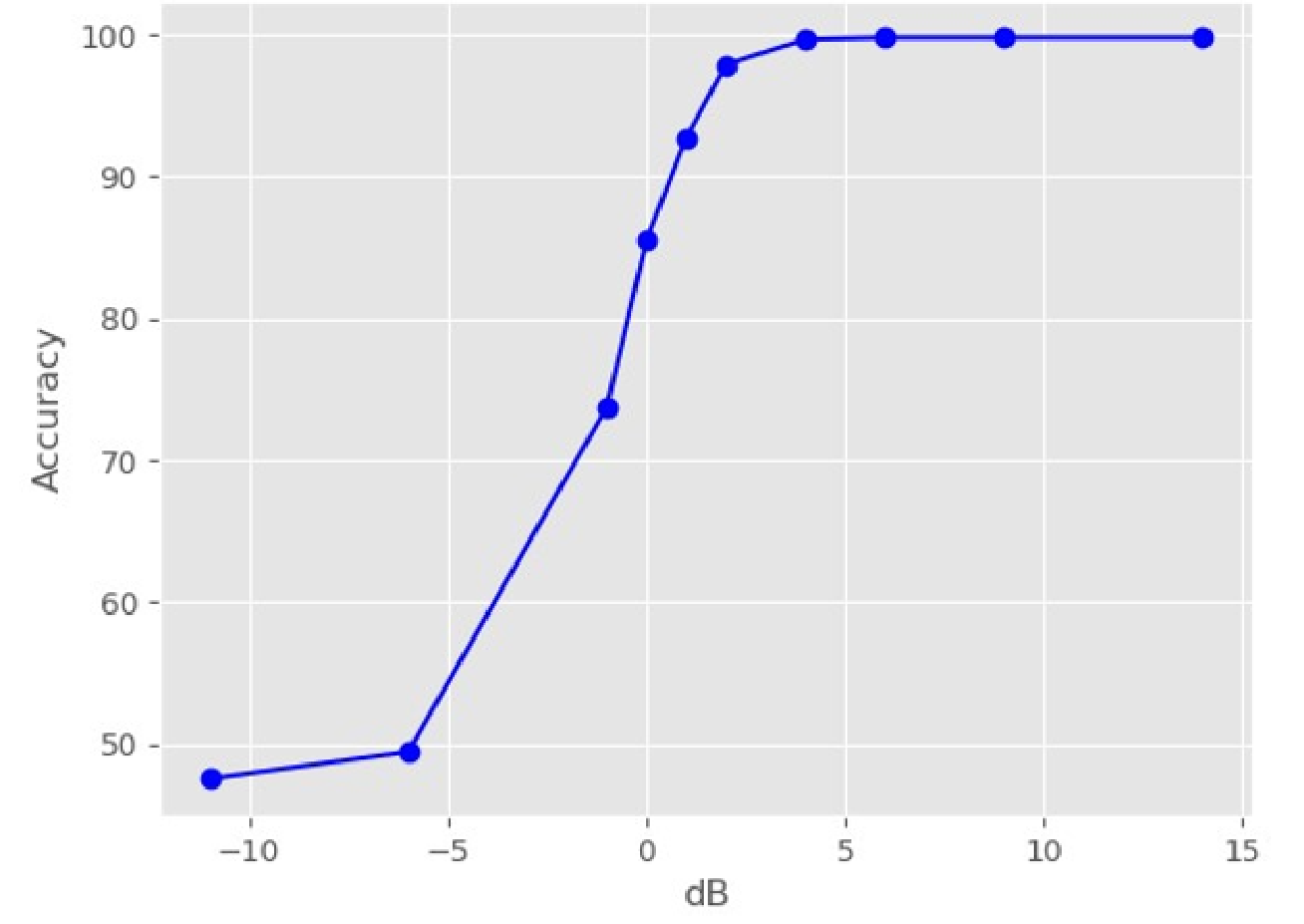}
  \caption{Accuracy versus SNR for EDA on spaceborne AIS data.}
  \label{eda_accuracy_plot}
\end{figure}

To assess the impact of noise on the EDA performance, we synthetically add additive white Gaussian noise (AWGN) to the real RF captures from space. The outcome, accuracy versus SNR, is shown in Figure \ref{eda_accuracy_plot} for the BCNN-based EDA model. The typical trend can be seen, where accuracy declines as SNR decreases. It is interesting to note that relying on decoding AIS burst (i.e. associate AIS bursts based on the extracted MMSI)  typically requires SNRs $\gg 2$ dB whereas ML-based EDA can be accomplished at significantly lower SNRs \cite{wawrzaszek2018detection}.

\begin{figure}[!ht]
\centering
\includegraphics[width=8.5cm, height=9.5cm]{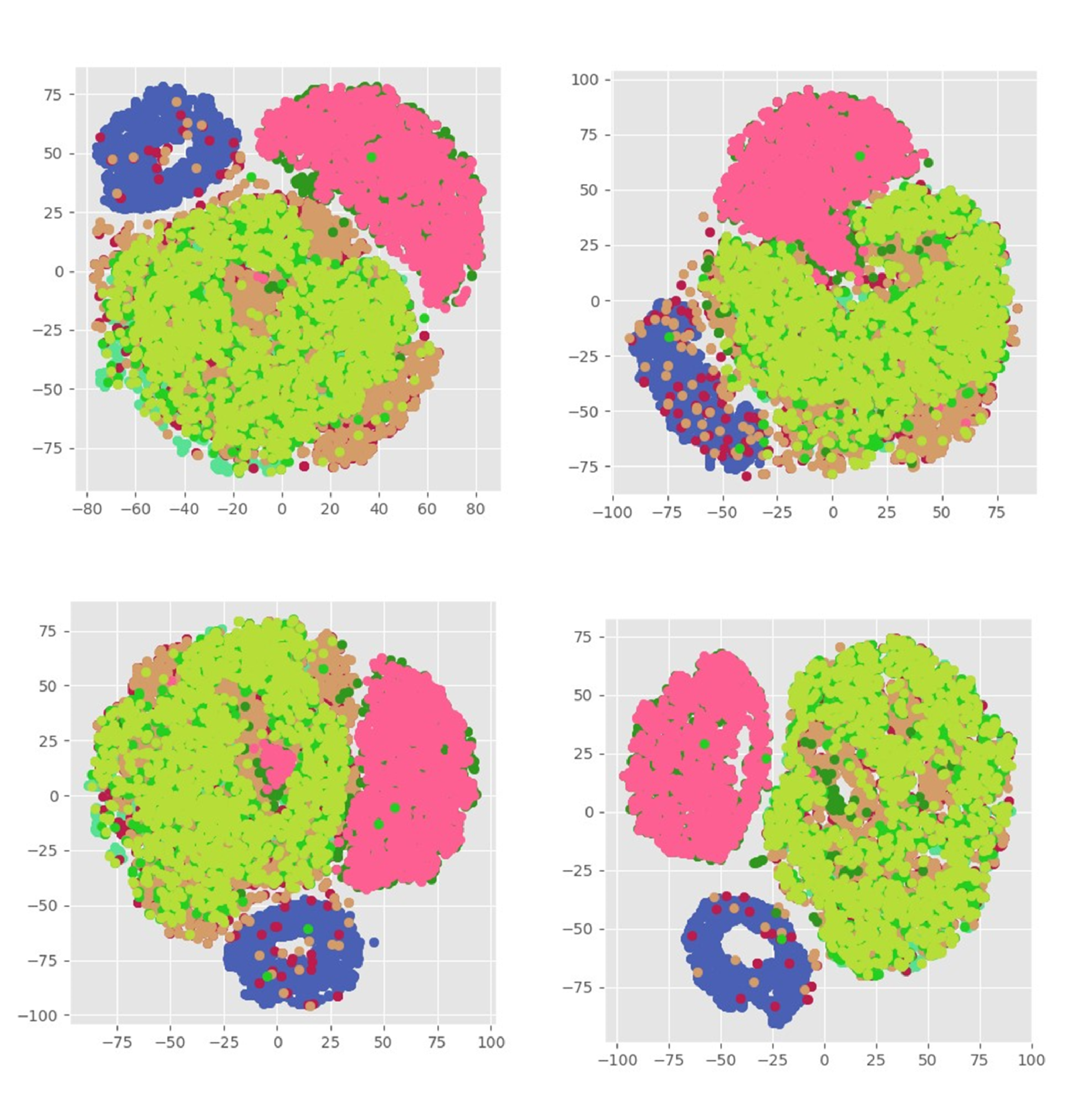}
\caption{t-SNE plot of DMR dataset embeddings following emitter representation learning step.
    Top: simpleAE (L), simpleconv1DAE (R).
    Bottom: vanillaAE (L), verysimpleAE (R).}
\label{bcnn_dist_eda}
\end{figure}
\begin{figure}[!t]
\centering
\includegraphics[width=8.5cm, height=9.5cm]{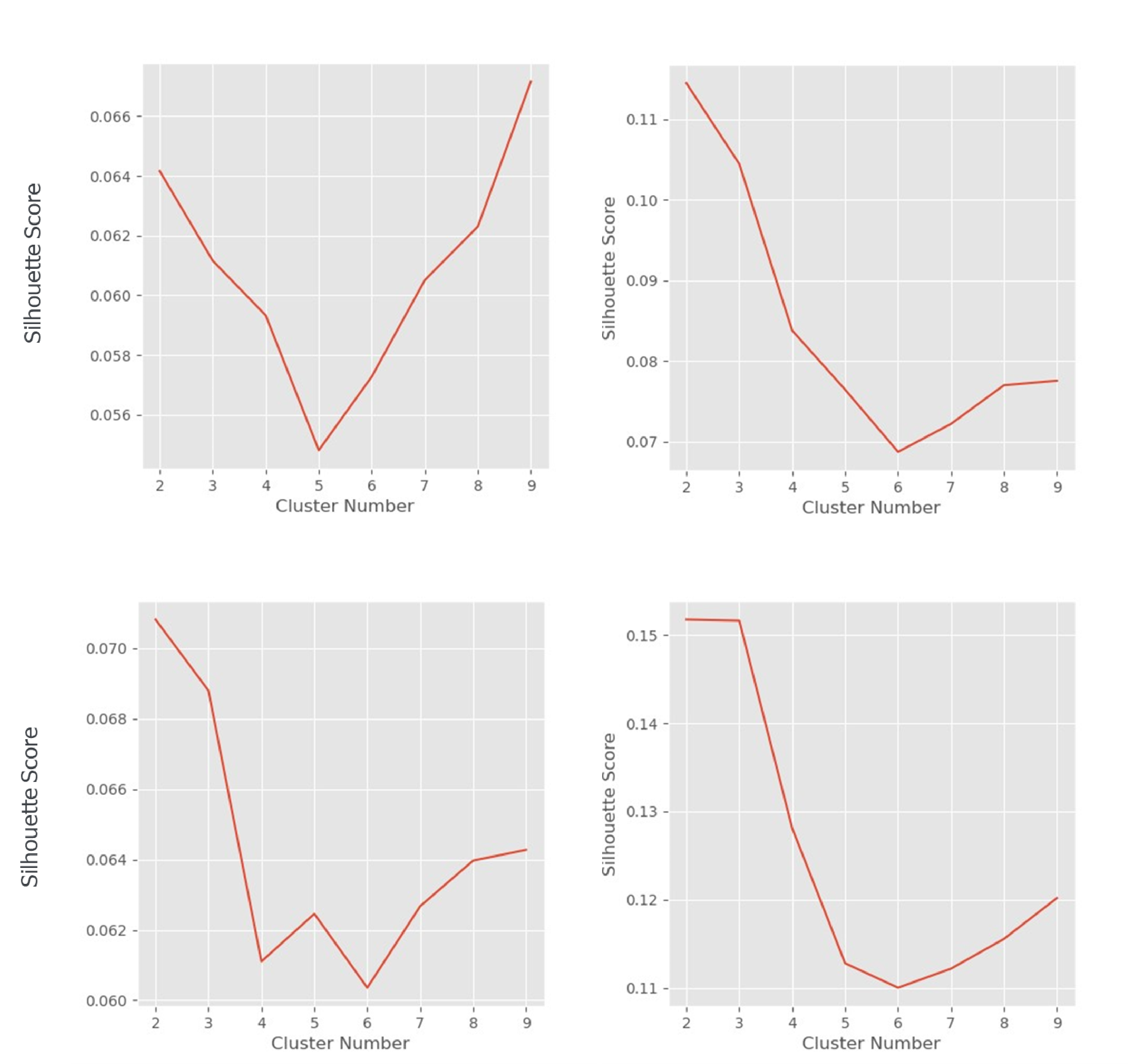}
\caption{K-Means Silhouette scores of DMR dataset embeddings.
        Top: simpleAE (L), simpleconv1DAE (R).
        Bottom: vanillaAE (L), verysimpleAE (R).}
\label{bcnn_dist_eda2}
\end{figure}
\subsection{Use Case 5: RFEC for DMR Data}\label{sec:RFEC_DMR}
Figure \ref{bcnn_dist_eda} depicts the t-SNE plots of the embeddings (RF fingerprint representation) for four model architectures using the DMR dataset. Notice how data structures have formed from learning lower dimensional representations of the emitter with autoencoders. Figure \ref{bcnn_dist_eda2} shows silhouette scores \cite{rousseeuw1987silhouettes}, which were computed on the training data embeddings, across a range of cluster numbers. It provides inconclusive evidence of emitters clusters being formed following unsupervised training. Nevertheless, it is included to demonstrate the RFEC task. It is possible that the DMR data has complexities which cannot be resolved via unsupervised approaches.

\begin{figure}[!b]
    \centering
    \includegraphics[width=5cm, height=4.1cm]{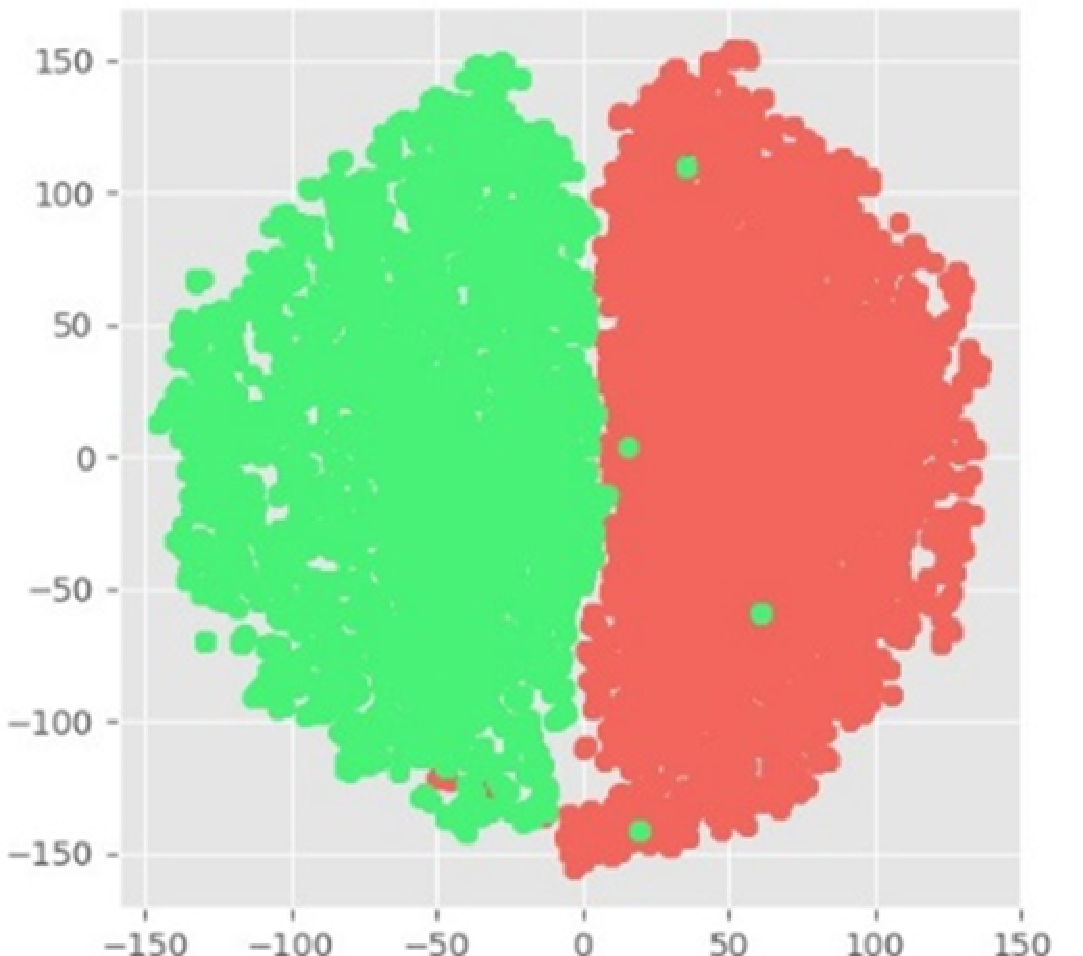}
    \includegraphics[width=5cm, height=4.1cm]{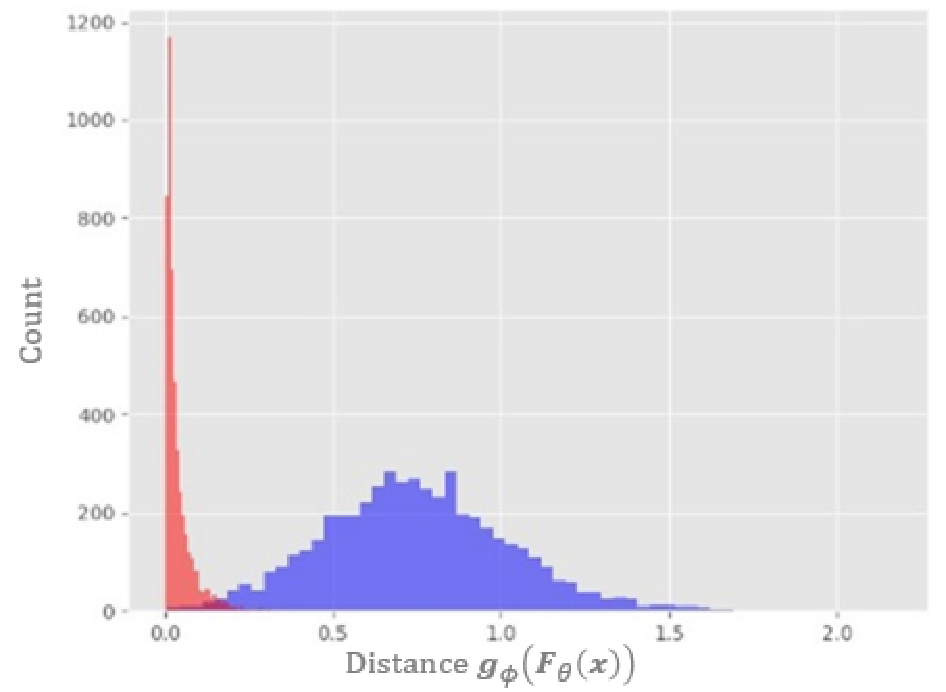}
    \includegraphics[width=5cm, height=4.1cm]{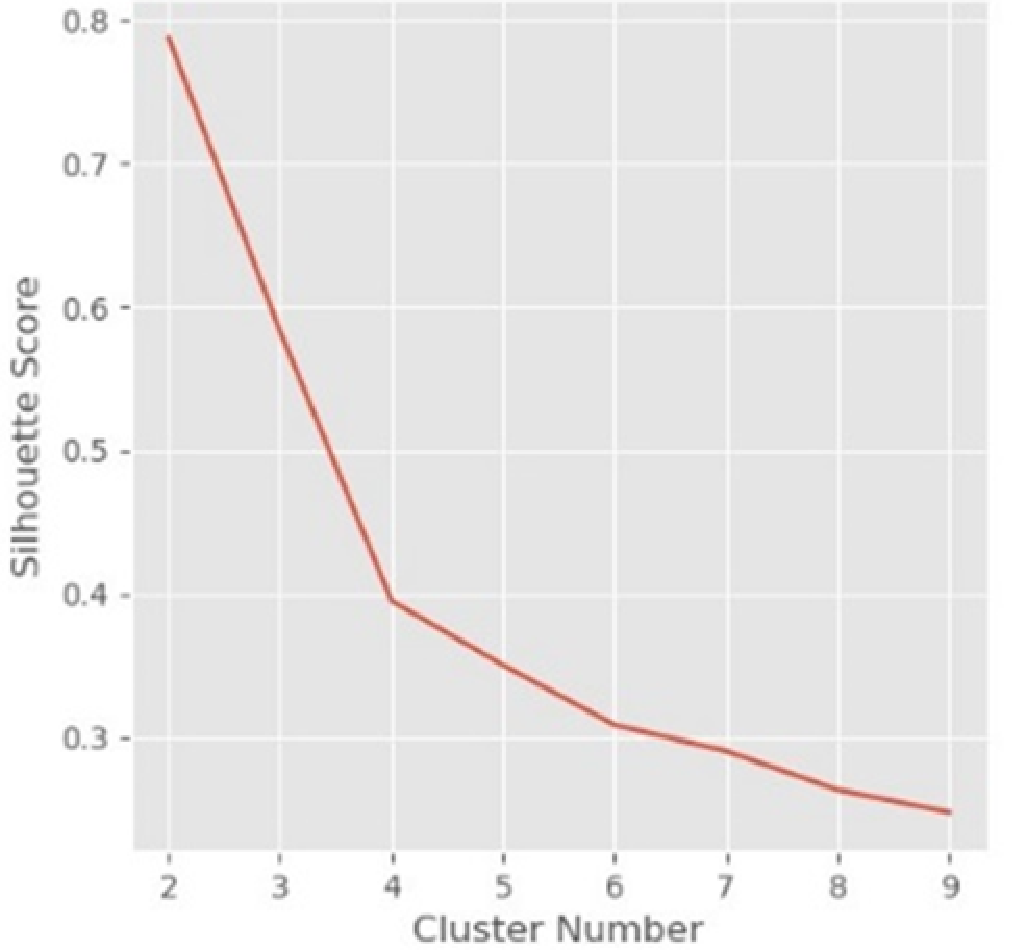}
    \caption{DMR OS-EDA results $p=0.5$. Top: BCNN t-SNE embeddings, matched (red) and unmatched (blue) input. Middle: pair distance distributions. Bottom: K-Means Silhouette score.}
    \label{eda_dmr_open_set}
\end{figure}

\begin{figure}[!t]
  \centering 
    \includegraphics[width=5cm, height=4.1cm]{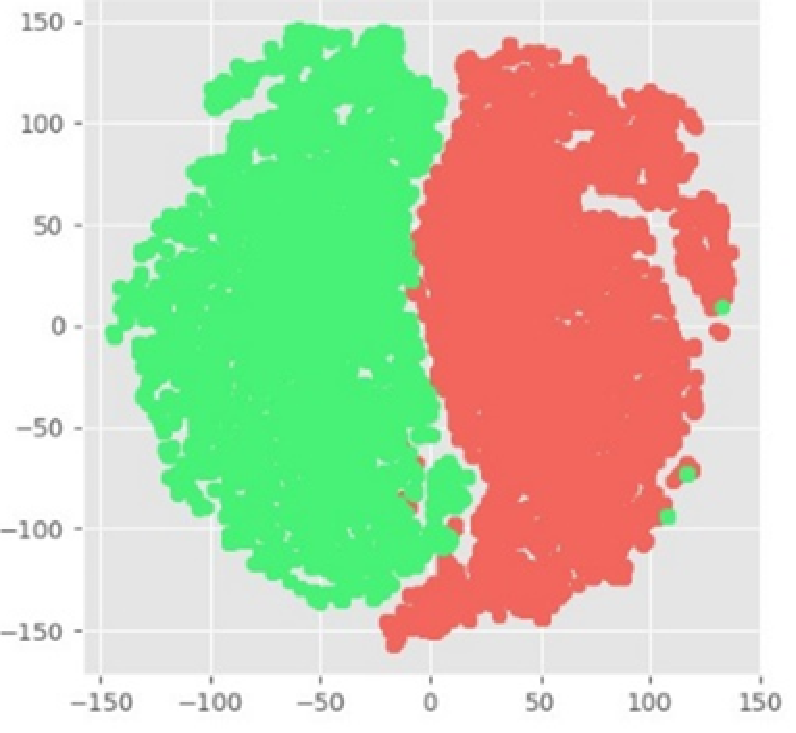}
    \includegraphics[width=5cm, height=4.1cm]{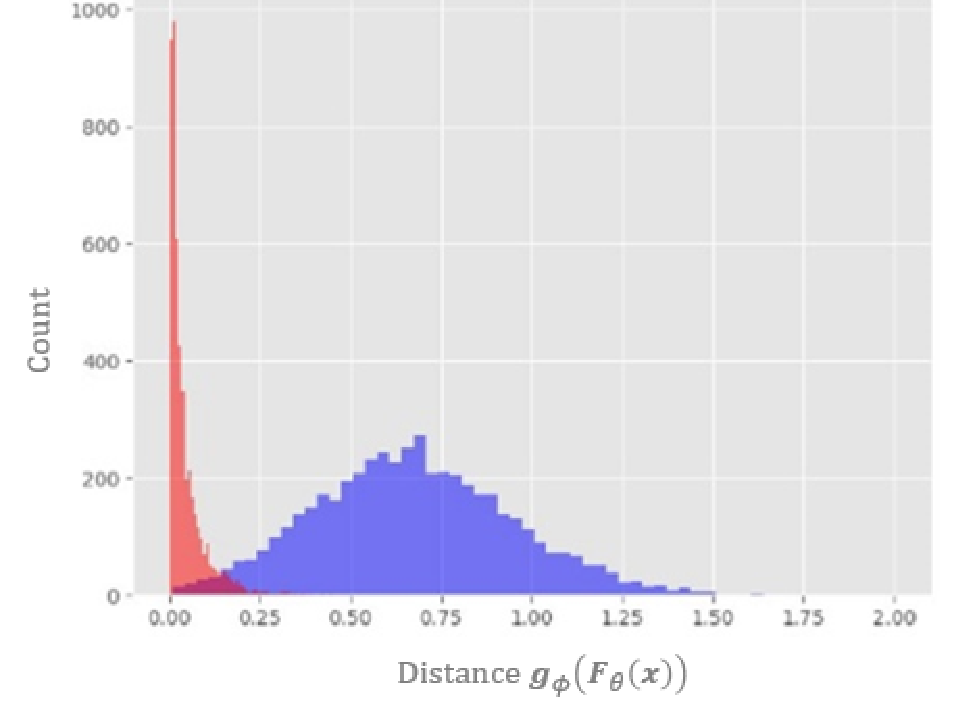}
    \includegraphics[width=5cm, height=4.1cm]{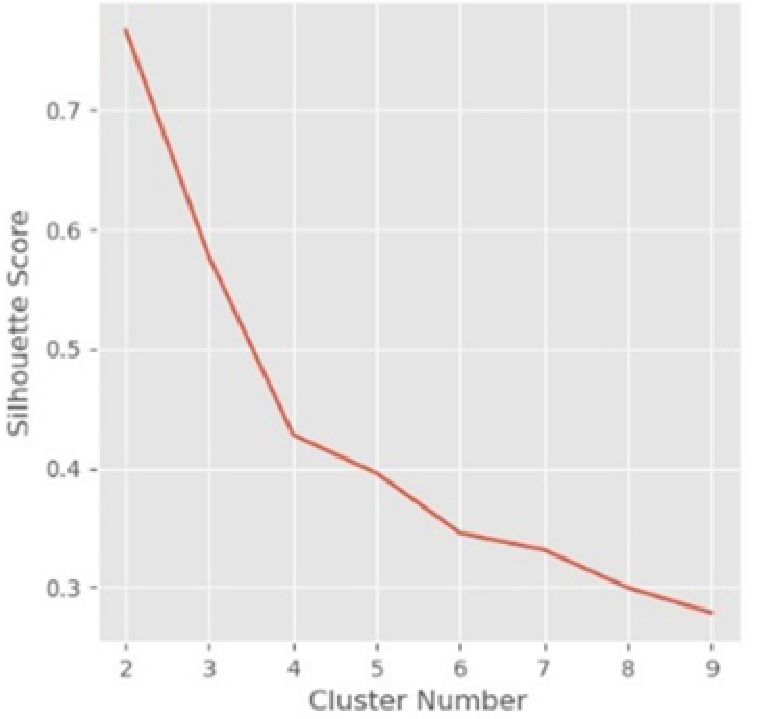}
    \caption{DMR OS-EDA results $p=0.25$. Top: BCNN t-SNE embeddings, matched (red) and unmatched (blue) input. Middle: pair distance distributions. Bottom: K-Means Silhouette score.}
    \label{eda_dmr_open_set1}
\end{figure}

\begin{figure}[!t]
\centering
    \includegraphics[width=5cm, height=4.1cm]{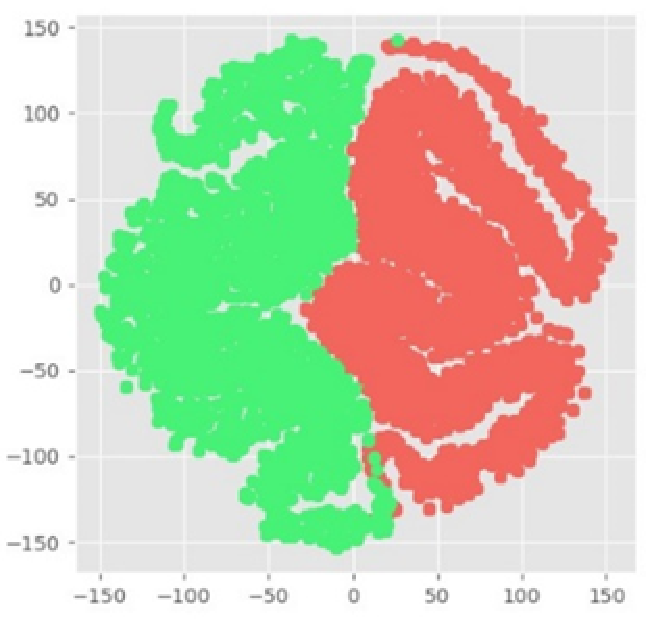}
    \includegraphics[width=5cm, height=4.1cm]{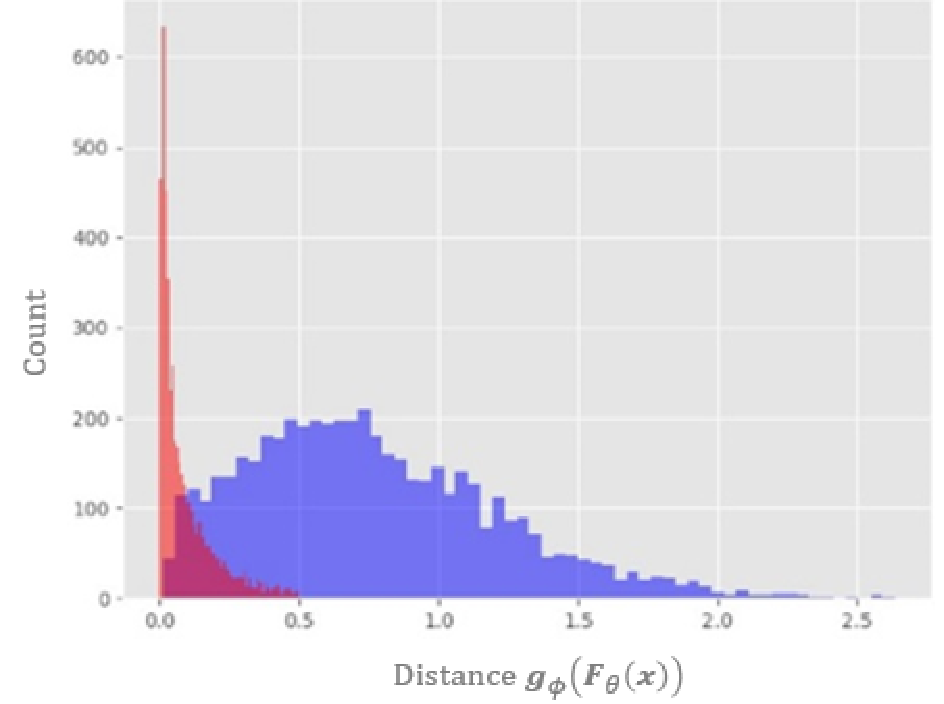}
    \includegraphics[width=5cm, height=4.1cm]{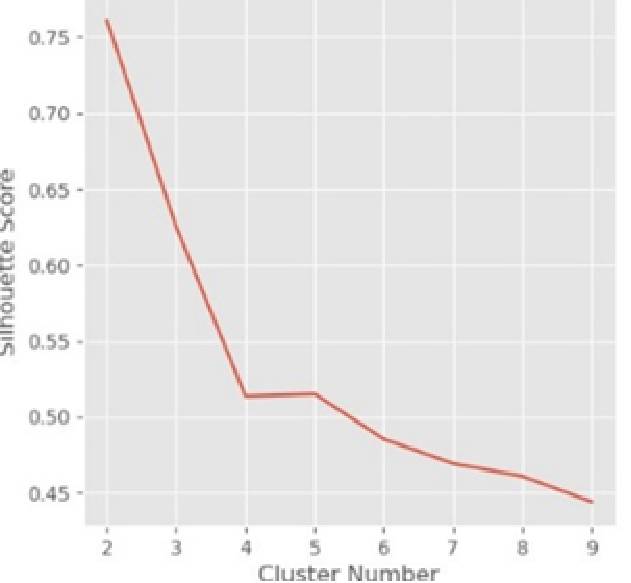}
    \caption{DMR OS-EDA results $p=0.025$. Top: BCNN t-SNE embeddings, matched (red) and unmatched (blue) input. Middle: pair distance distributions. Bottom: K-Means Silhouette score.}
    \label{eda_dmr_open_set2}
\end{figure}

\subsection{Use Case 6: OS-EDA for DMR data} \label{sec:OSEDA_DMR}
We present here open-set EDA task results with the DMR dataset. An EDA model can compare input pairs without their emitter labels, meaning we can train an EDA model on known emitters and evaluate it on
potentially unseen emitters, hence OS-EDA. This is similar to the methodology in Section \ref{sec:singletask_OS}, however for EDA. We trained an EDA model on the DMR dataset using emitter data from six handsets. We held out the data of the remaining two emitters' to be used for testing to assess the model's OS capability.  In addition, we adjusted a proportion parameter $p$ of the training data to understand how much data would be needed to obtain sufficiently good performance (i.e. $p=1$ uses all of training dataset, $p=0.5$ uses half of it, etc) on the two held out emitters' data. 

Figures \ref{eda_dmr_open_set}, \ref{eda_dmr_open_set1} and \ref{eda_dmr_open_set2} display a summary of the evaluation on data for held out emitters with a BCNN architecture for EDA with $p=0.5, 0.25, 0.025$ respectively. It can be noted from all the three figures that the t-SNE embeddings of each emitter are distinguishable. Whilst the distance distributions have some overlap between them, overall they are well separated. The silhouette score is also high for cluster number 2, giving more indication on the structure of the evaluated (two) emitters' data. As $p$ is reduced, the performance marginally deteriorates. These are promising experimental results on training generic RFF models. 

\section{Conclusions}\label{sec:conclusions}
This paper introduces a framework for building a general-purpose machine learning RF fingerprinting model. Within this, we describe a variety of RF fingerprint-dependent downstream tasks and demonstrate the performance of a number of them, namely SEI, EDA, RFEC and OS-EDA, using three real RF datasets (i.e. DMR for SIGINT, AIS for spaceborne surveillance and drone RF signals for C-UAS).  From this, we observe that low complexity ML model architectures deliver a competitive performance when compared with their more complex counterparts across all tasks and datasets. This supports the implementation of high performance data-driven RFF techniques on edge devices, which are typically characterised by a low size, weight, power and cost hardware profile. The presented generic ML-based framework for RF fingerprinting and presented results can serve as an impetus to future research on data-driven RFF.  

\appendices
\section{Data Transformation Operator for EDA}
We introduce in this Appendix a sampling algorithm $\mathcal{M}$ for the EDA task. We assume that $\mathcal{D}^{(a)} = \mathcal{M}(\mathcal{D}^{(i)}) = \{ ( {\bf{x}}_j^{(a),1}, {\bf{x}}_j^{(a), 2}, y^{(a)}_j) \}_{j=1}^{n_{a}}$ where we choose 
$n_{a} = \gamma \ll \frac{n_i(n_i-1)}{2}$, $\gamma > 0$. Setting this condition on the number of samples in the EDA dataset is to avoid the combinatorial explosion of pairs when $n_i$ is large. This way, we can sample from all possible pairs without completion, which could 
lead to a large computational overhead for some problems. For EDA labeling, we have: 
\begin{equation}
    \tilde{y}(i, j) =\begin{cases}
      1, & \text{if $y_i=y_j$},\\
      0, & \text{otherwise},
    \end{cases}
  \end{equation}
where $y_i, y_j \in \{1, \cdots, \mathcal{V} \}$. In order to create one data point ${\bf{z}}\in \mathcal{D}^{(a)}$, we sample twice from $\mathcal{D}^{(i)}$: $({\bf{x}}_1, y_1) \sim \mathcal{D}^{(i)}, ({\bf{x}}_2, y_2) \sim \mathcal{D}^{(i)} \backslash ({\bf{x}}_1, y_1)$. 
Following the definition of $\tilde{y}(i, j)$, a data point $d = (x_1, x_2, \tilde{y}(1, 2))$ is created.

\begin{algorithm}[!t]
  \caption{Dataset sampling for EDA}
  \label{alg:example}
\begin{algorithmic}
  \State {\bfseries INPUTS:} proportion $\alpha$, number of EDA samples $\gamma$, total number of classes $\mathcal{V}$, dataset $\mathcal{D}^{(i)}$
  \State {\bfseries INITIALIZE:} $\mathcal{D}^{(a)} = \{ \}$
  \For{$j=1$ {\bfseries to} $\mathcal{V}$}
  \For{$k=1$ {\bfseries to} $\mathcal{V}$}
  \State $T = \mathcal{C}_{jk}(\mathcal{V}; \alpha, \gamma)$
  \For{$v=1$ {\bfseries to} $T$}
  \State Sample $d_1 = ({\bf{x}}_1, j) \sim \mathcal{D}^{(i)}_j$ and 
  \State Sample $d_2 = ({\bf{x}}_2, k) \sim \mathcal{D}^{(i)}_k \backslash d_1$
  \State Set $d = ( {\bf{x}}_1, {\bf{x}}_2, \tilde{y}(j, k) )$
  \State Add $d$ to $\mathcal{D}^{(a)}$ 
  \EndFor
  \EndFor
  \EndFor
  \State {\bfseries OUTPUT:} EDA dataset $\mathcal{D}^{(a)}$ 
\end{algorithmic}
\end{algorithm}

In general, the EDA problem has class imbalance because as the number of classes $\mathcal{V}$ increases, the ratio of matched to unmatched pairs declines. The number of matched $n_{matched}= \mathcal{V}$ and the number of unmatched $n_{unmatched}= \frac{\mathcal{V}(\mathcal{V}-1)}{2}$. 
This allows us to define the ratio $n_r = \frac{n_{matched}}{n_{unmatched}}=\frac{\mathcal{V}}{ \frac{1}{2}\mathcal{V}(\mathcal{V}-1)}=\frac{2}{\mathcal{V}-1}$. 
This can inform how to structure the sampling, in terms of proportionality. We now define a function:
\begin{equation}
    \mathcal{C}_{ij}(\mathcal{V}; \alpha, \gamma) =\begin{cases}
      \frac{\alpha\gamma}{n}, & \text{if $i=j$},\\
      \frac{2(1-\alpha)\gamma}{n(n-1)}, & \text{if $i \neq j$};
    \end{cases}
  \end{equation}
which determines the number of input pairs for every class pairing $(i, j)$, $i, j \in \{1, \cdots, \mathcal{V}\}$, that is to be included in ${\mathcal{D}}^{(a)}$. 
The parameter $\alpha \in (0, 1]$ is a ratio of matched to unmatched pairs and $\gamma$ is the total number of samples that is sampled from $\mathcal{D}^{(i)}$. 
We can structure $\mathcal{D}^{(i)}$ such that 
\begin{equation}
\mathcal{D}^{(i)} = \{ \mathcal{D}^{(i)}_1, \cdots, \mathcal{D}^{(i)}_n  \} \; \mbox{where} \; \mathcal{D}^{(i)}_k = \{ ({\bf{x}}_j^{(i)}, k) \}_{j=1}^{n_i^k}
\end{equation}
where $n_i^k$ is the number of samples in $\mathcal{D}^{(i)}_k$. Algorithm \ref{alg:example} shows a recipe for computing dataset $\mathcal{D}^{(a)}$ using $\mathcal{D}^{(i)}$.

\bibliographystyle{IEEEtran}
\bibliography{IEEEfull,paper}
\end{document}